%% file: iclr2026_conference.tex
\documentclass{article} 
\usepackage{iclr2026_conference,times}

\input{math_commands.tex}

\usepackage{array}
\usepackage{hyperref}
\usepackage{url}
\usepackage{subcaption}
\usepackage{multicol}
\usepackage{multirow}
\usepackage{booktabs}
\usepackage{fontawesome5}
\usepackage{xcolor}
\usepackage{academicons}
\usepackage{fontawesome5}
\usepackage{fontawesome5}
\definecolor{myblue}{HTML}{1F77B4}
\definecolor{myred}{HTML}{D62728}
\usepackage{graphicx}
\title{Extreme Region Policy Distillation}

\author{Changyu Chen$^{1}$, Xiting Wang$^{1}$, Rui Yan$^{2}$ \\
$^{1}$Gaoling School of Artificial Intelligence, Renmin University of China\\
$^{2}$Wuhan University \\
chen.changyu@ruc.edu.cn
}

\iclrfinalcopy 
\begin{document}

\maketitle

\begin{abstract}
Reinforcement learning for large language models faces a fundamental trade-off between sample efficiency and asymptotic performance: strictly on-policy methods discard trajectories after a single update, while off-policy reuse introduces distribution mismatch that existing trust-region techniques mitigate primarily by enforcing conservative optimization, often leaving rich training signals underutilized. To investigate this, we perform extensive off-policy updates on fixed data. Our experiments reveal that aggressive multi-step optimization brings rapid initial gains, but excessive updates cause trajectory probabilities to deviate and entropy to collapse, with performance plateauing early. Tightening KL constraints merely lowers the ceiling without resolving the degradation. This motivates Extreme Region Policy Distillation (ERPD), a two-stage framework that decouples sample efficiency from KL efficiency. The first stage performs weakly constrained off-policy optimization on fixed data to maximally extract training signals. The resulting policy provides token-level supervision. In the second stage, we distill these signals into the base policy under trust-region constraints, filtering harmful drift while preserving useful signals. The distilled policy achieves comparable or better performance with substantially smaller KL divergence, indicating that much of the first-stage divergence was spent on unnecessary drift rather than genuine improvement. Crucially, ERPD accommodates both strong and weak teachers: when aggressive optimization yields no stronger policy, even degenerate teachers provide effective supervision via alternative signal construction strategies. We validate ERPD on mathematical reasoning, showing gains for strong base models where on-policy training plateaus, and reliable improvements with weak teachers.
\end{abstract}

\begin{center}
\begin{tabular}{cc}
\href{https://huggingface.co/adalaw/Qwen3.5-27B-ERPD-003}{\includegraphics[width=1em]{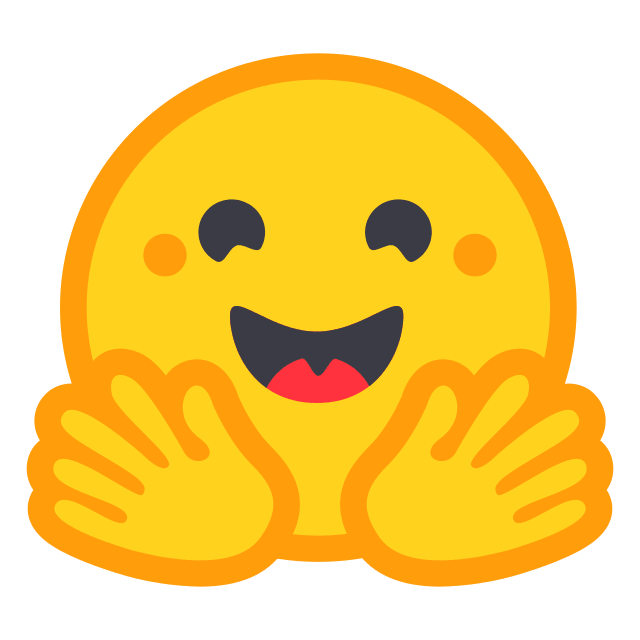} HuggingFace} & 
\href{https://github.com/ChangyuChen347/ERPD}{\faGithub\ Github}
\end{tabular}
\end{center}

\input{iclr2026/src/intro}

\input{iclr2026/src/method}
\input{iclr2026/src/exp}

\input{iclr2026/src/related}

\bibliography{iclr2026_conference}
\bibliographystyle{iclr2026_conference}


\end{document}

%% file: math_commands.tex
\usepackage{amsmath,amsfonts,bm}

\def\eqref#1{equation~\ref{#1}}

\def\1{\bm{1}}

\DeclareMathAlphabet{\mathsfit}{\encodingdefault}{\sfdefault}{m}{sl}
\SetMathAlphabet{\mathsfit}{bold}{\encodingdefault}{\sfdefault}{bx}{n}



%% file: iclr2026/src/intro.tex
\section{Introduction}

\begin{figure}[ht]
    \centering
    \begin{subfigure}[b]{0.66\linewidth}
        \centering
        \includegraphics[width=\linewidth]{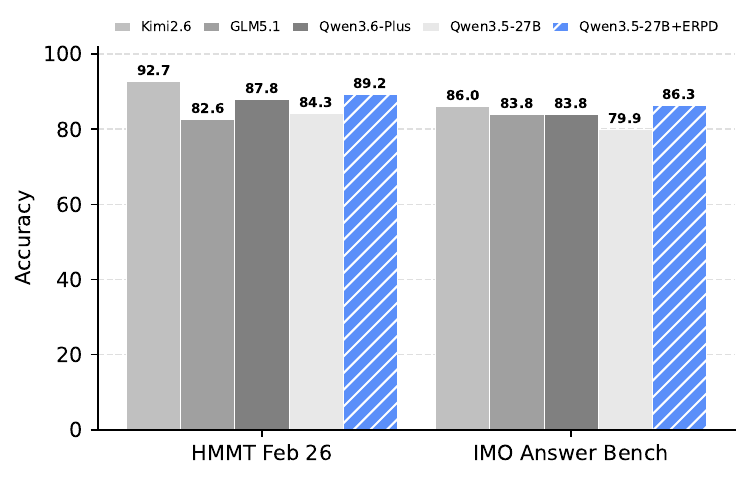}
        \caption{Qwen3.5-27B trained with ERPD matches frontier-level models.}
        \label{fig:1a}
    \end{subfigure}
    \hfill
    \begin{subfigure}[b]{0.31\linewidth}
        \centering
        \includegraphics[width=\linewidth]{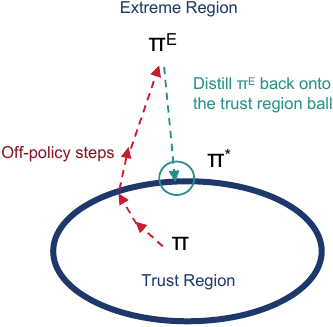}
        \caption{The proposed ERPD.}
        \label{fig:gain}
    \end{subfigure}
    \caption{(a) Demonstration of performance gains. (b) The framework decouples aggressive off-policy teacher optimization from KL-constrained student distillation.}
    \label{fig:overall}
\end{figure}

In recent years, reinforcement learning (RL) has driven substantial advances in the reasoning capabilities of large language models, yielding breakthroughs in mathematical problem-solving and code generation~\citep{ouyang2022training,guo2025deepseek}. 
Unlike supervised learning, which optimizes toward fixed targets via direct imitation, RL discovers novel reasoning strategies through trial-and-error interaction guided by reward signals~\citep{mroueh2025reinforcement}. 
However, this flexibility comes at a cost: generating long-horizon trajectories incurs significant computational overhead, and the strictly on-policy nature of these updates means collected data is discarded after a single gradient step. While off-policy methods can improve sample efficiency by reusing trajectories, repeatedly optimizing on the same batch gradually shifts the policy away from the behavior distribution that generated the data. Such distribution mismatch destabilizes training, revealing a fundamental trade-off between data efficiency and optimization stability~\citep{xi2025bapo}.

Existing stabilization techniques operate within this trade-off, constraining policy updates through likelihood-ratio clipping or KL divergence penalties~\citep{ppo,trpo} to prevent destabilization.
Yet in practice, these constraints enforce a conservative optimization paradigm: even when a batch contains rich, underutilized training signals, it is typically discarded after a few gradient updates. 
This conservatism raises a critical question: how much potential improvement is lost by not fully exploiting each batch? 
To investigate this, we examine the effect of performing extensive off-policy updates on fixed trajectory data. 
Our experiments reveal a clear pattern: aggressive multi-step optimization brings rapid initial gains at first, but too many updates cause trajectory probabilities and policy entropy to drift apart. Final performance therefore plateaus early, suggesting that much of the accumulated KL divergence comes from unnecessary policy drift rather than real task improvement. Such wasted deviation consumes the limited KL budget without actually helping performance, leaving little room for later, more effective updates. Furthermore, simply tightening KL constraints for off-policy updates does not improve KL efficiency; it only lowers the performance ceiling by being too conservative.

This observation motivates the central objective of this paper: can we retain the benefits of aggressive off-policy optimization while eliminating its ineffective KL consumption? 
We address this challenge through a distillation perspective. 
Specifically, we propose a two-stage training framework that explicitly decouples sample efficiency from KL efficiency. 
\begin{figure}[h]
    \centering
    \begin{subfigure}[t]{0.27\linewidth}
        \centering
        \includegraphics[width=\linewidth]{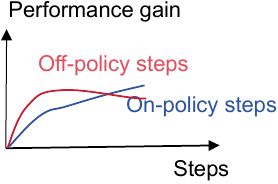}
        \caption{Previous framework requires a trade-off.}
    \end{subfigure}
    \hfill
    \begin{subfigure}[t]{0.65\linewidth}
        \centering
        \includegraphics[width=\linewidth]{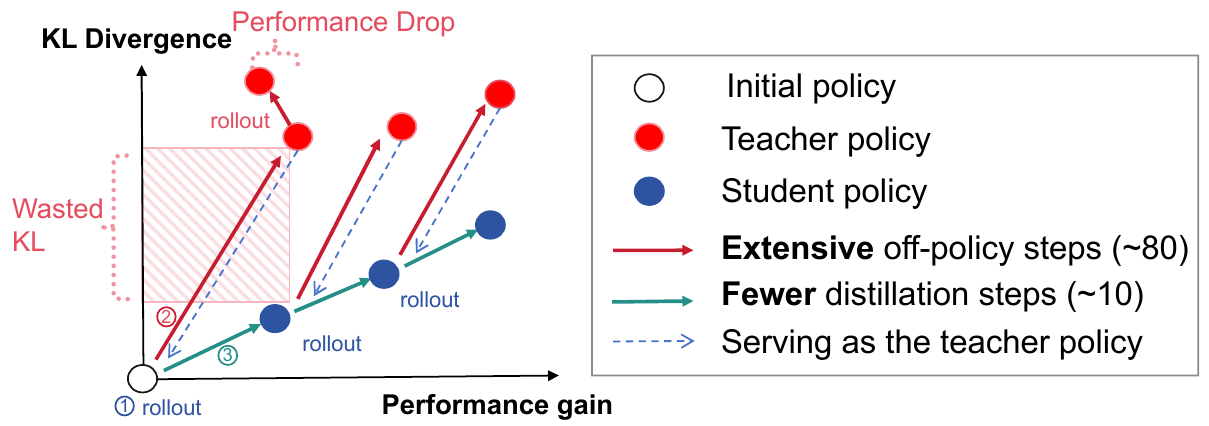}
        \caption{Our two-stage method decouples sample efficiency from KL efficiency, optimizing each separately.}
         \label{fig:2b}
    \end{subfigure}
    \caption{The proposed ERPD framework resolves the trade-off between sample efficiency and asymptotic performance via decoupled optimization. (a) Conventional methods must compromise between data reuse and optimization stability. (b) Our two-stage approach separates aggressive off-policy signal extraction from trust-region constrained distillation, enabling both high sample efficiency and strong final performance.}
     \label{fig:2}
\end{figure}
In the first stage, we prioritize sample utilization by relaxing KL constraints, enabling extensive off-policy optimization on a fixed dataset to extract maximal training signal---even at the cost of larger policy divergence. 
The resulting policy serves not as the final model but as a teacher that provides rich supervisory signals. 
In the second stage, we distill these signals into the original policy under explicit trust-region constraints, preserving the effective information discovered in the first stage while filtering out the associated spurious drift. 
Fig.~\ref{fig:2b} illustrates the resulting policy achieves comparable performance with substantially smaller KL divergence, enabling more sustained improvement.

\begin{figure}[h]
    \centering
    \includegraphics[width=0.95\linewidth]{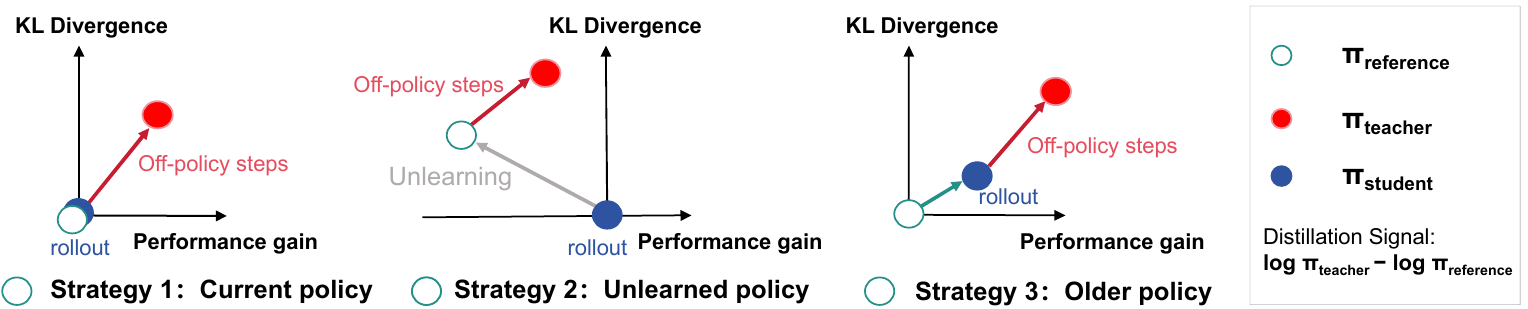}
    \caption{In our experiments, we primarily adopt the first strategy; however, it does not always succeed in producing a stronger teacher. When that happens, the second and third strategies provide effective signals.}
    \label{fig:1}
\end{figure}

However, obtaining a stronger teacher policy at each round of off-policy iteration is not guaranteed.
We further show that even a more extreme form of update — using a degenerate, weaker teacher — can still provide useful supervisory signals when a stronger teacher cannot be trained.
Specifically, we analyze different strategies for constructing these distillation signals (Fig.~\ref{fig:1}). The primary approach, Strategy 1, uses the current policy as a reference model. When off-policy updates under Strategy 1 fail to produce a stronger teacher, we find that switching to Strategy 2 or 3 for signal construction can lead to better performance. The results in Fig.~\ref{fig:1a}, obtained with Strategy 2 on Qwen3.5-27B, further demonstrate the potential of these alternative strategies to improve state-of-the-art models even in the absence of a strong teacher.

We summarize our contributions as follows:
\begin{enumerate}
    \item We introduce a two-stage training framework that first performs aggressive, weakly constrained policy optimization and then distills the resulting signals back into the base policy under explicit KL constraints, achieving improved sample efficiency and KL efficiency.
    \item For the strong teacher strategy, we compare the sample efficiency of various policy losses, verify that distillation reduces KL divergence, and analyze the conditions under which the student can outperform the strong teacher.
    \item For the weak teacher strategy, we provide a systematic empirical characterization, verify the feasibility of weak-to-strong distillation, and present detailed ablations.
    \item We demonstrate on mathematical reasoning tasks that, for strong base models where standard on-policy optimization yields diminishing returns, our method delivers substantial performance gains.
\end{enumerate}

%% file: iclr2026/src/method.tex
\section{Preliminary}

\subsection{Notations}

Let $\theta_{\text{old}}$ denote the parameters of the old policy, and $\theta$ the parameters of the candidate policy. Define the probability ratio
$
r(\theta) = \frac{\pi_{\theta}(a \mid s)}{\pi_{\theta_{\text{old}}}(a \mid s)}.
$
Denote the advantage function estimated under the old policy by $A^{\pi_{\theta_{\text{old}}}}(s,a)$, and let $\rho_{\pi_{\theta_{\text{old}}}}$ be the state-visitation distribution induced by $\pi_{\theta_{\text{old}}}$.

\subsection{Token-Level Training Signals}
Improving the sample efficiency of LLM training is a central challenge in reinforcement learning–based alignment~\citep{ouyang2022training,guo2025deepseek}. One effective avenue toward this goal is the use of dense reward signals, particularly at the token level, which provide fine-grained supervision compared to sparse, sequence-level feedback. Token-level rewards enable learning algorithms to extract substantially more information from each rollout, accelerating convergence and reducing the need for expensive data collection~\citep{RTO}.
Recent methods have demonstrated that such dense rewards can be automatically constructed~\citep{rafailov2024r,ce}, without relying on external labeling. A prominent example is Direct Preference Optimization (DPO)~\citep{rafailov2023direct}, which transforms pairwise preference data into token-level log-ratio rewards relative to a reference policy.   
%
\citet{rafailov2024r} further point out that, under the maximum-entropy RL framework, the policy learned by DPO implicitly induces a token-level reward function.
Specifically, the reward for taking action $a$ in state $s$ is given by
\begin{equation}
R(s,a) 
= \beta \log \frac{\pi_\theta(a \mid s)}{\pi_{\theta_{\text{old}}}(a \mid s)},
\label{eq:rr}
\end{equation}
where $\beta$ is a hyperparameter in DPO training that controls the deviation of the candidate policy $\pi_\theta$ from the reference policy (commonly understood as the strength of KL-divergence regularization).

Building on this insight, \citet{ce} further propose to directly optimize the reward defined in Eq.~\ref{eq:rr} using a cross-entropy (CE) loss, thereby eliminating the need for pairwise preference comparisons required by DPO. 
Given a sequence $y=(a_1,\ldots,a_T)$ and a binary preference label $l \in \{0,1\}$, the loss is defined as
\begin{equation}
L_{\text{CE}} 
= l \cdot \log \sigma\big(R(y)\big) 
+ (1-l) \cdot \log \left[1 - \sigma\big(R(y)\big)\right],
\end{equation}
where the sequence-level reward decomposes into the sum of token-level rewards:
\begin{equation}
R(y) 
= \sum_{t=1}^{T} R(s_t,a_t)
= \beta \sum_{t=1}^{T} \log \frac{\pi_\theta(a_t \mid s_t)}{\pi_{\theta_{\text{old}}}(a_t \mid s_t)}.
\end{equation}

However, this formulation gives rise to a set of underexplored questions: if both the reward model (teacher) and the target policy (student) share the same initialization and training data, What is the value of distillation with the token-level reward? Specifically, can the student surpass the teacher, and are there other benefits beyond performance? We will analyze these questions empirically in the experimental sections.

\subsection{Trust Region Methods}

Standard policy gradient methods adhere strictly to the on-policy training paradigm: trajectories sampled under the current policy are typically used for only one parameter update, resulting in low sample efficiency. A natural approach to improving data utilization is to reuse trajectories generated by an older policy $\pi_{\theta_{\text{old}}}$ when updating the current policy $\pi_\theta$, which introduces importance sampling.
However, when the new policy deviates too far from the old one, importance weights can become unstable, leading to high variance in gradient estimates and unreliable updates~\citep{espeholt2018impala,roux2025tapered}. Consequently, if one wishes to reuse data while maintaining stable performance improvement, it is necessary to explicitly limit the magnitude of policy change between successive updates.

Trust region methods address this issue by constraining policy updates to remain in the vicinity of the previous policy. A common practice is to measure the discrepancy between policies using the KL divergence and enforce it to remain below a preset threshold. This idea underlies Trust Region Policy Optimization (TRPO), which aims to achieve stable and reliable policy improvement while permitting limited reuse of data.

The constrained optimization formulation of TRPO is
\begin{equation}
\begin{aligned}
\theta_{\text{new}} &= \arg\max_{\theta} \;
\mathbb{E}_{s \sim \rho_{\pi_{\theta_{\text{old}}}}, a \sim \pi_{\theta_{\text{old}}}} 
\Bigl[ r(\theta) \, A^{\pi_{\theta_{\text{old}}}}(s,a) \Bigr] \\
\text{s.t.} \quad &
\mathbb{E}_{s \sim \rho_{\pi_{\theta_{\text{old}}}}} 
\Bigl[ D_{\text{KL}}\bigl( \pi_{\theta_{\text{old}}}(\cdot|s) \,\big\|\, \pi_{\theta}(\cdot|s) \bigr) \Bigr] \le \delta,
\end{aligned}
\label{eq:trpo}
\end{equation}
where $\delta>0$ denotes the trust-region radius, which limits the magnitude of policy updates.

Because TRPO is complex to implement, Proximal Policy Optimization (PPO) proposes a more practical approximation that constrains the policy update by clipping the probability ratio, thereby suppressing both excessive gains and excessive penalties, which indirectly limits policy deviation~\citep{ppo}. This loss design has been widely adopted in the RL training of large language models, with representative methods such as Group Relative Policy Optimization (GRPO) following a similar clipping principle~\citep{grpo}.

The clipped surrogate objective of PPO:
\begin{equation}
\begin{aligned}
\theta_{\text{new}} = \arg\max_{\theta} \,
\mathbb{E}_{
\substack{s \sim \rho_{\pi_{\theta_{\text{old}}}} \\ a \sim \pi_{\theta_{\text{old}}}}
}
\Bigl[ \min\!\bigl( r(\theta) A^{\pi_{\theta_{\text{old}}}}(s,a),\,  
\operatorname{clip}\!\bigl(r(\theta), 1-\varepsilon, 1+\varepsilon\bigr) A^{\pi_{\theta_{\text{old}}}}(s,a) \bigr) \Bigr],
\end{aligned}
\label{eq:ppo}
\end{equation}
where $\varepsilon$ is a hyperparameter controlling the clipping range. The clipping operation sets the gradient to zero whenever $r(\theta)$ exceeds $[1-\varepsilon,\,1+\varepsilon]$, effectively capping the magnitude of policy updates.

\begin{figure*}[t]
    \centering
    \begin{subfigure}{0.32\textwidth}
        \centering
        \includegraphics[width=\linewidth]{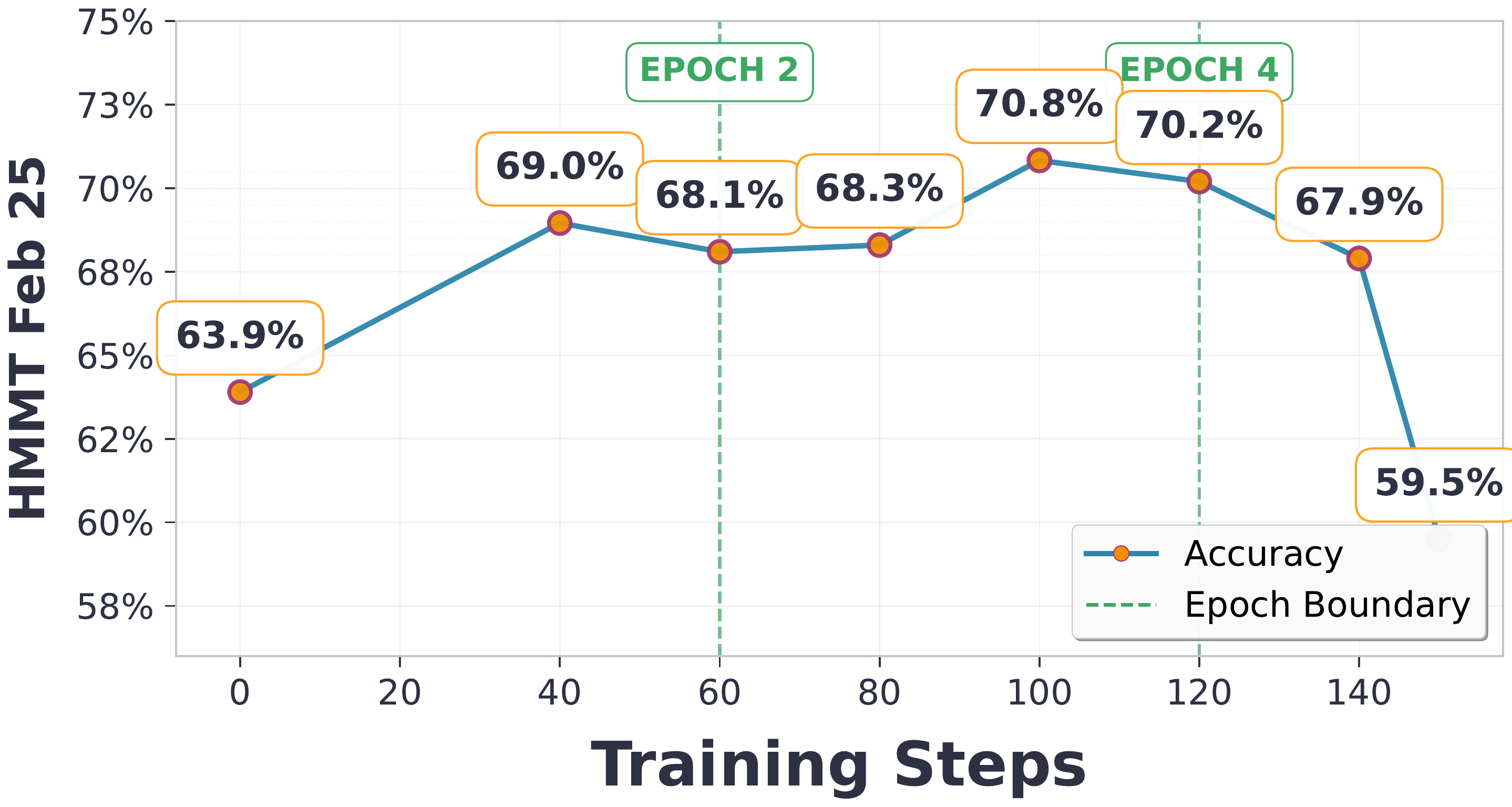}
        \caption{Training NBG4-3B with SAPO.}
        \label{subfig:fig1}  
    \end{subfigure}
    \hfill
    \begin{subfigure}{0.32\textwidth}
        \centering
        \includegraphics[width=\linewidth]{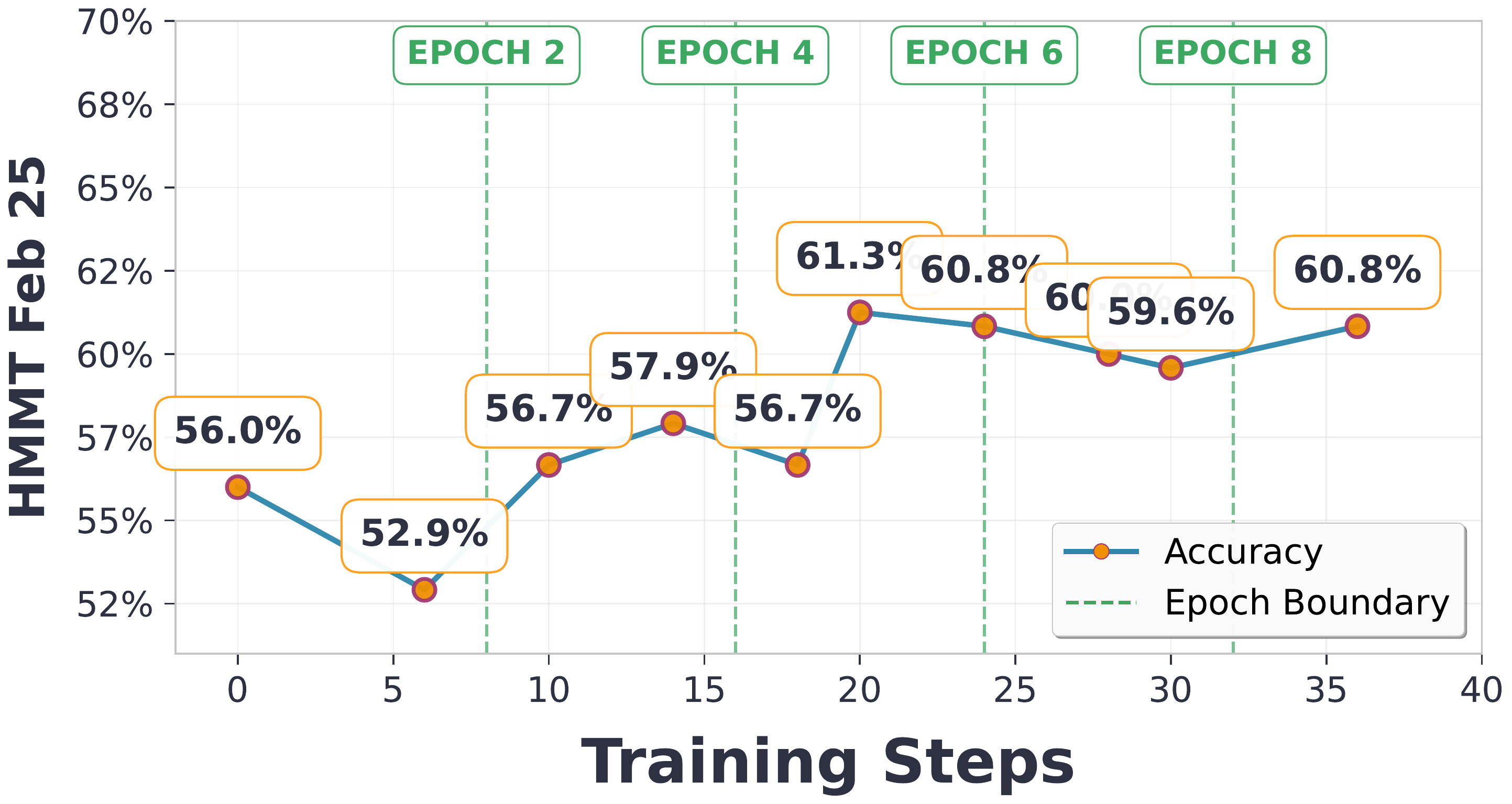}
        \caption{Training Qwen3-4B with CE.}
        \label{subfig:fig2}
    \end{subfigure}
    \hfill
    \begin{subfigure}{0.32\textwidth}
        \centering
        \includegraphics[width=\linewidth]{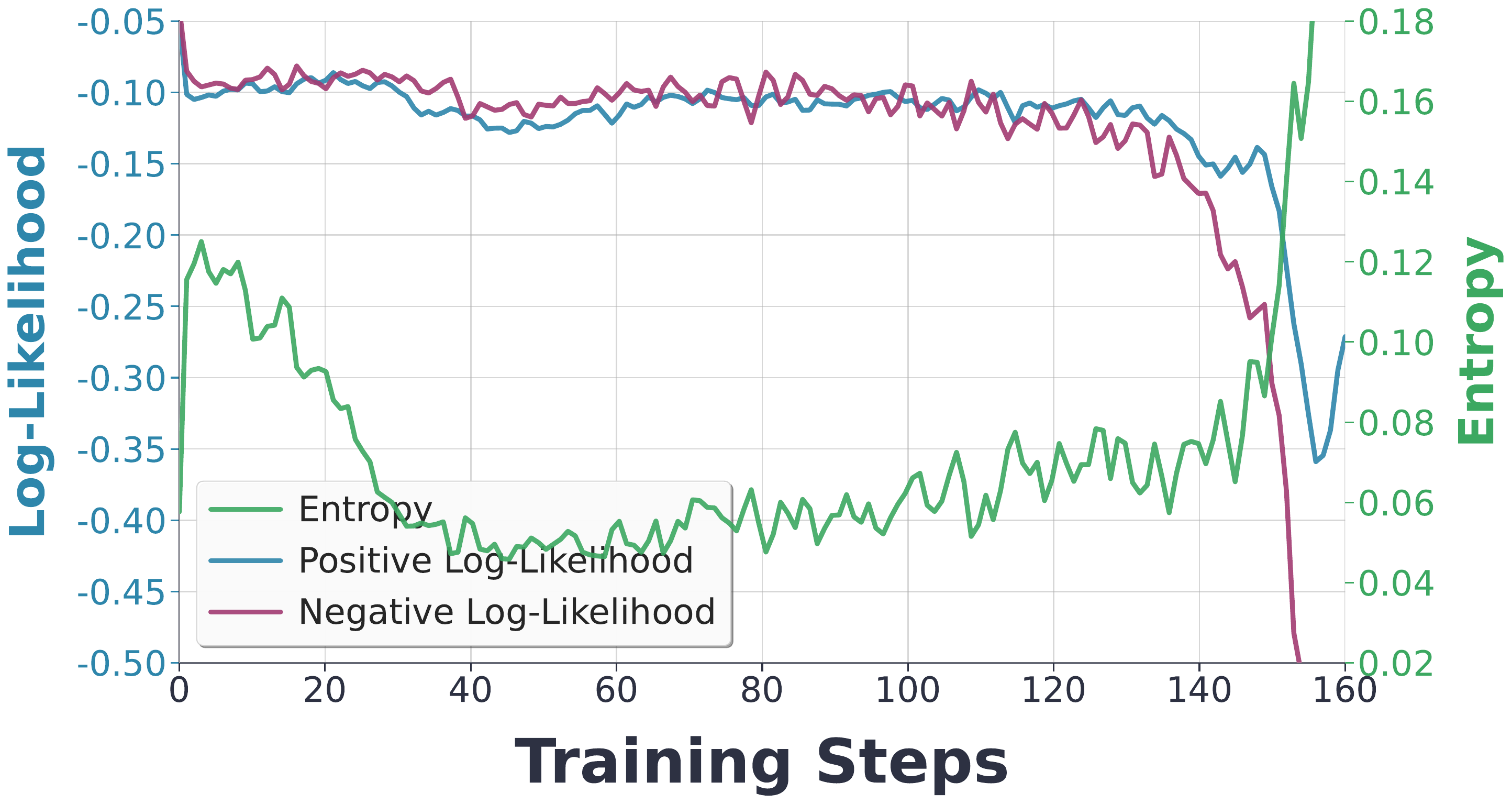}
        \caption{Instability in Optimization.}
        \label{subfig:fig4}
    \end{subfigure}
    \caption{The experiments are conducted on Qwen3-4B-2507-Thinking (Qwen3-4B) and Nanbeige4-3B-2511-Thinking (NBG4-3B), using a fixed batch of 1K prompts with corresponding rollouts. We perform multi-step off-policy updates on this fixed data. The loss function is SAPO~\citep{sapo} and CE~\citep{ce}, which replace hard clipping with a soft weighting mechanism, potentially making it less conservative compared to hard clipping~\citep{ppo}.}
    \label{fig:combined}
    \label{fig:pre}
\end{figure*}

\section{Decoupled Two-Stage Optimization}
Extreme Region Policy Distillation is a two-stage optimization framework. We introduce these two stages in Sec.~\ref{sec:sample} and Sec.~\ref{sec:kl}, respectively, present implementation details in Sec.~\ref{sec:imp}, and then, in Sec.~\ref{sec:mse}, describe how to construct weak-to-strong distillation signals from a degenerate teacher policy.


\subsection{Stage 1: Off-policy Updates for Sample Efficiency}
\label{sec:sample}
In current reinforcement learning frameworks for LLM reasoning, policy updates over collected rollouts are typically conservative.
Some methods strictly follow an on-policy regime and perform only a single update per rollout batch~\citep{he2025skyworkopenreasoner1},
while others apply multi-step optimization using minibatches drawn from the same batch.
Even in the latter case, the number of update steps is usually small, often around four~\citep{he2025justrlscaling15bllm,sapo}.

\textbf{Preliminary Experiments.}
Building on the question raised in the introduction regarding how much information remains underutilized due to conservative policy constraints, 
we examine how aggressively a fixed rollout batch can be optimized, and how many gradient updates are required to extract its learning signal.
Within a single iteration, we collect a fixed batch of rollouts using the old policy.
We then perform a large number of gradient update steps on this static dataset, producing an extreme region policy $\pi_{\theta_e}$.
Throughout this process, we track how policy performance evolves as the number of optimization steps increases. We show the result in Fig.~\ref{fig:pre}.

\textbf{Observation 1 (Conservative Updates Underutilize Rollout Batches).}
When optimizing for sample efficiency, performing only a small number of update steps (e.g., four or fewer) remains overly conservative.
As shown in Figs.~\ref{subfig:fig1} and ~\ref{subfig:fig2}, a fixed batch of rollouts typically requires dozens of optimization steps before its learning signal is fully exploited.  6-step optimization even leads to a performance drop in Fig.~\ref{subfig:fig2}.
This suggests that standard training protocols leave significant performance gains unrealized.

\textbf{Observation 2 (Policy Drift under Long-Horizon Optimization).}
As the number of update steps increases, prolonged optimization on a fixed batch inevitably induces substantial policy drift.
Specifically, large changes in token-level log-probabilities and entropy are observed, causing the updated policy to deviate significantly from the old policy (Fig.~\ref{subfig:fig4}).
In the later stages of training, this issue is further exacerbated by gradient imbalance, which can cause a sharp decline in the probabilities of both positive and negative examples, ultimately degrading training stability.

Taken together, these observations reveal a fundamental trade-off: aggressive optimization is necessary to fully exploit rollout batches and improve sample efficiency, yet it inevitably pushes the policy into extreme regions of the policy space.
In the following sections, we show how the learning signals extracted from such extreme region policies can be effectively distilled into a stable, constrained policy.

\subsection{Stage 2: Distillation for KL-Efficiency}
\label{sec:kl}
We now turn to the distillation method used in the second stage, which stabilizes these extreme-region policies while preserving their performance gains.
Under a limited KL divergence budget, an ideal policy update should allocate the deviation primarily to directions that yield genuine performance improvement. In practice, however, structural biases in the loss function and distributional shift can cause the policy to incur additional KL divergence along directions that are only weakly related to performance gains. As a result, a considerable portion of the KL divergence budget is spent on directions that contribute little to the objective.

We focus on improving what we term KL divergence efficiency—that is, the amount of policy improvement achieved per unit of KL divergence.
Ideally, if we could explicitly identify the directions of policy change that are unrelated to improvement (or equivalently, construct a policy $\pi_{\theta_-}$ that is orthogonal to improvement), we could then formulate the following constrained optimization problem:
\begin{equation}
\min_{\pi}
\;\mathrm{KL}(\pi \,\|\, \pi_{\theta_e})
\quad
\text{s.t.}
\quad
\mathrm{KL}(\pi \,\|\, \pi_{\theta_-}) \ge \epsilon.
\label{eq:ge}
\end{equation}
In general, however, such a decomposition is difficult to achieve because $\pi_{\theta_-}$ is not easy to define, making a practical alternative necessary.

\textbf{Trust Region Constrained Distillation.}
To arrive at a practical alternative, we return to the idea of trust regions: the policy update is confined to a region where it remains reliable.

Specifically, we approximately solve the following constrained optimization problem:
\begin{equation}
\min_{\pi}
\;\mathrm{KL}(\pi \,\|\, \pi_{\theta_e})
\quad
\text{s.t.}
\quad
\mathrm{KL}(\pi \,\|\, \pi_{\theta_{\text{old}}}) \le \epsilon,
\label{eq:le}
\end{equation}
where $\pi_{\theta_{\text{old}}}$ is the reference policy that generated the rollout batch.
This formulation can be interpreted as a distillation process: while keeping the policy within the vicinity of $\pi_{\theta_{\text{old}}}$ (the trust-region constraint), it draws the policy as close as possible to $\pi_{\theta_e}$, thereby distilling the components of $\pi_{\theta_e}$ that contribute to performance improvement.

Intuitively, one may understand distillation as improving KL divergence efficiency in the following sense.
The TRPO lemma bounds the policy optimization error by the KL divergence~\citep{trpo}:
\begin{equation}
|\text{Err}(\pi)| = |L_{\pi_{\text{old}}}(\pi) - \eta(\pi)| \leq C \cdot D_{\text{KL}}(\pi_{\text{old}} \parallel \pi).
\end{equation}
$L_{\pi_{\text{old}}}(\pi)$ is the off-policy surrogate objective defined in TRPO. Optimizing this surrogate objective $L$ introduces an error with respect to the true performance $\eta(\pi)$. Consequently, the closer the updated policy stays to $\pi_{\text{old}}$, the tighter the bound on the optimization error incurred by policy deviation.

\subsection{Implementation}
\label{sec:imp}
\textbf{Token Reward Signal.} Minimizing $\mathrm{KL}(\pi_\theta \,\|\, \pi_{\theta_e})$ is equivalent, up to a sign, to maximizing the expected log-ratio
$\mathbb{E}_{\pi_\theta}[\log \pi_{\theta_e} - \log \pi_\theta]$.
The resulting gradient takes a policy-gradient-like form, involving $\nabla_\theta \log \pi_\theta$ weighted by a log-probability difference between the current and extreme policies.
In practice, we approximate the constrained optimization in Eq.~\ref{eq:le} using importance weighted samples from $\pi_{\theta_{\text{old}}}$ and optimize a PPO-style clipped surrogate.

We define a reward signal function as
$
\hat{A}(s,a)
=
\log \frac{\pi_{\theta_e}(a \mid s)}{\pi_{\theta_{\text{old}}}(a \mid s)}.
$
Here we use $\pi_{\theta_{\text{old}}}$ in the denominator (instead of the evolving $\pi_\theta$) so that $\hat{A}$ remains fixed during optimization. Our experiments suggest that this choice reflects a trade-off. As $\pi_\theta$ is pulled closer to $\pi_{\theta_e}$, the influence of  
$
\log \frac{\pi_{\theta_e}(a \mid s)}{\pi_\theta(a \mid s)}
$
gradually diminishes, whereas  
$
\log \frac{\pi_{\theta_e}(a \mid s)}{\pi_{\theta_{\text{old}}}(a \mid s)}
$
remains fixed throughout training. The former is therefore more KL-friendly, while the latter is empirically more likely to achieve performance that surpasses $\pi_{\theta_e}$.
Due to the scale sensitivity of the log-ratio, we apply whitening normalization to $\hat{A}$.

\textbf{Clipped Surrogate Objective.}
Using $\hat{A}(s,a)$ to replace the advantage signal, we optimize the following PPO-style objective:
\begin{equation}
\label{eq:stage2}
\theta_{\text{new}} = \arg\max_{\theta} \,
\mathbb{E}_{
\substack{s \sim \rho_{\pi_{\theta_{\text{old}}}} \\ a \sim \pi_{\theta_{\text{old}}}}
}
\Bigl[ \min\!\bigl( r(\theta)\hat{A}(s,a),\,  
\operatorname{clip}\!\bigl(r(\theta), 1-\varepsilon, 1+\varepsilon\bigr)\hat{A}(s,a) \bigr) \Bigr],
\end{equation}
where $r(\theta)=\frac{\pi_{\theta}(a \mid s)}{\pi_{\theta_{\text{old}}}(a \mid s)}$.

\subsection{Distilling from Unlearned Extreme Region Policy}
\label{sec:mse}
In contrast to the extreme region policies considered in Sec.~\ref{sec:sample}, which are trained under weak KL regularization, we now study a fully unconstrained optimization regime.
Specifically, we remove the KL constraint to the old policy and optimize the policy loss until action probabilities become highly saturated toward 0 or 1.


We train the policy using a simple mean squared error (MSE) objective:
\begin{equation}
\mathcal{L}_{\text{MSE}}
=
\mathbb{E}_{s \sim \rho_{\pi_{\theta_{\text{old}}}}, a \sim \pi_{\theta_{\text{old}}}} 
\left[
\bigl(\pi_{\theta}(a \mid s) - R\bigr)^2
\right],
\end{equation}
where $R \in \{0,1\}$ denotes the terminal outcome reward of the trajectory.
This objective is analogous to fitting a Monte Carlo return with a critic, but parameterized directly through the policy head to induce probability saturation in the extreme region.

\begin{figure}[ht]
    \centering
    \includegraphics[width=0.7\linewidth]{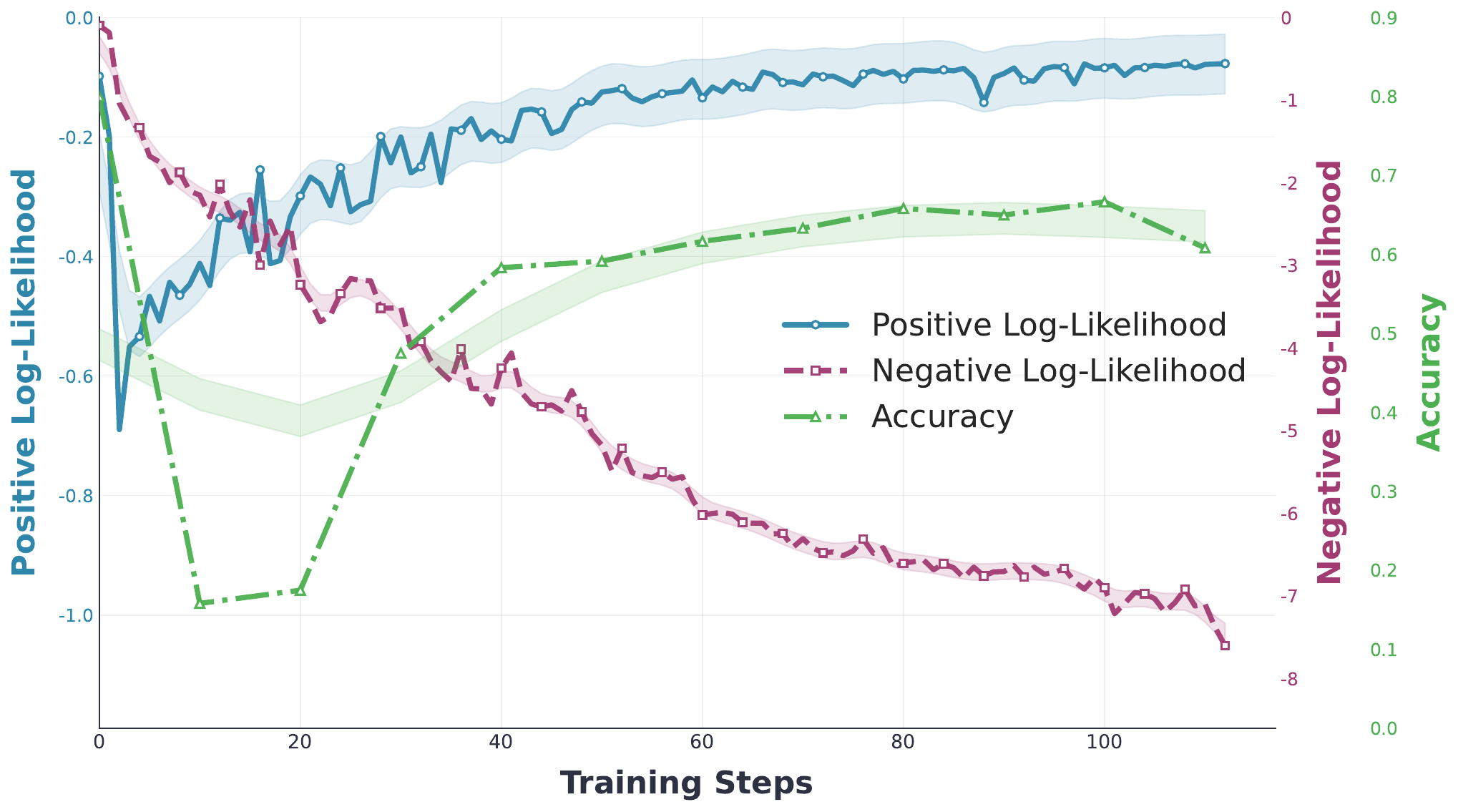}
    \caption{Typical training dynamics with MSE loss.}
    \label{fig:perf}
\end{figure}

\textbf{Training Dynamics.}
A typical optimization trajectory is shown in Fig.~\ref{fig:perf}.
During early training, the predicted probabilities for positive examples decrease rapidly, accompanied by a temporary drop in validation accuracy.
As optimization proceeds, the model gradually recovers, and both positive probability mass and accuracy increase.
Despite this unlearn-and-recover dynamics, the final policy can also achieve a near-zero MSE loss and an explained variance exceeding 0.9, indicating a fitting capacity comparable to that of a standard critic with a linear head.


\textbf{Token Reward Signals.} 
Our preliminary experiments show that using
$
\hat{A}(s,a) = \log \frac{\pi_{\theta_e}(a \mid s)}{\pi_{\theta_{\text{old}}}(a \mid s)}
$
does not lead to performance improvements in this setting. Through empirical exploration, we find that an alternative construction is more effective. Specifically, we define the reward signal as
$
\hat{A}(s,a)
=
\log \frac{\pi_{\theta_e}(a \mid s)}{\pi_{\theta_{\text{un}}}(a \mid s)}.
$
Here, $\pi_{\theta_{\text{un}}}$ (short for $\pi_{\theta_{\text{unlearned}}}$) denotes an intermediate policy checkpoint obtained during early optimization, rather than the old policy.
Empirically, during the initial phase of training, gradients from negative examples dominate, causing the model to temporarily suppress probabilities for positive examples, a process we refer to as \emph{unlearning}.
We select $\pi_{\theta_{\text{un}}}$ from this early \textbf{unlearning phase} (typically between steps 10 and 30), after which the model enters a \textbf{recovery phase} as shown in Fig.~\ref{fig:perf}.

%% file: iclr2026/src/exp.tex
\section{Experiments}
\subsection{Settings}
\textbf{Compared Methods.}
We use the notation ``X+Distillation'' to refer to our two-stage  pipeline, where ``X'' denotes the loss used to train the teacher and ``+Distillation'' indicates distillation from the teacher.

For the teacher loss functions, we compare GRPO~\citep{mroueh2025reinforcement}, PPO~\citep{ppo}, and a soft clipping method called SAPO~\citep{sapo}. In addition, we introduce Cross Entropy (CE)~\citep{ce,cui2025process} for comparison. CE shares a similar optimization approach with DPO but does not require paired positive and negative examples. Instead, it parameterizes $\log\frac{\pi_{\theta}(a|s)}{\pi_{old}(a|s)}$ as a reward model and fits the reward using a cross-entropy loss.
It is worth noting that both CE and SAPO leverage the soft clipping capability of the sigmoid function. We also compare against our proposed MSE loss. Detailed experiments about MSE will be presented in Sec.~\ref{sec:expmse}.


For online methods, we compare on-policy methods in Sec.~\ref{sec:on}, as well as iterations of our proposed two-stage approach.

\textbf{Models.} In our experiments, we select high-performance language models as starting points to ensure meaningful comparisons of sample efficiency. Specifically, we use Qwen3-4B-2507-Thinking, Nanbeige4-3B-2511-Thinking~\citep{yang2025nanbeige4}, Qwen3.5-9B, and Qwen3.5-27B. These models span a relatively wide range of parameter sizes, represent recent architectures, and already exhibit strong baseline performance, making them instructive testbeds for studying sample efficiency. For brevity, we refer to Qwen3-4B-2507-Thinking as Qwen3-4B and Nanbeige4-3B-2511-Thinking as NBG4-3B.

\textbf{Benchmarks.} We evaluate on mathematical reasoning tasks, including AIME24, AIME25, HMMT Feb 25, HMMT Nov 25, HMMT Feb 26~\citep{balunovic2025matharena}, IMO Answer Bench~\citep{luong2025robustmathematicalreasoning} and Beyond AIME~\citep{guo2025seed1}. Since sampling with temperature introduces large variance, we adopt the AVG@K metric~\citep{guo2025deepseek} for aggregation, setting K=16 for Beyond AIME, K=4 for IMO Answer Bench, and K=32 for the other datasets. Unless otherwise specified, the maximum sequence length is set to 81,920 for Qwen3-4B, 65,536 for NBG4-3B, and 192,000 for Qwen3.5 models.
We also conduct experiments on coding task and test on LiveCodeBench~\citep{jain2024livecodebench}.

Additionally, some tables report the KL divergence. All KL values in the tables are reverse KL divergence unless otherwise specified, computed as:

$$
\text{KL}[\pi_\theta \| \pi_{\theta_{\text{old}}}] \approx \frac{1}{N} \sum_{i=1}^{N} \left[ \log \pi_\theta(a_i \mid s_i) - \log \pi_{\theta_{\text{old}}}(a_i \mid s_i) \right]
$$

where $(s_i, a_i)$ are samples drawn from the current new policy $\pi_\theta$.

\textbf{Training Settings.}
We follow the data collection of POLARIS~\citep{Polaris2025}, which consists of challenging math problems and their corresponding answers.

In the offline setting, we first sample responses for 1,000 prompts to construct a static dataset. The default sampling parameters are as follows: temperature is set to 0.6 (or 1.0 for Qwen3.5 models), with 16 trajectories sampled per prompt, and a default response length of 32,768 — this response length may be extended based on the model’s generation capacity. Refer to the provided code scripts for configuration details. Under this setup, this offline data collection process is equivalent to the data collection step for optimizing a single batch in online training, with the distinction that our study performs more optimization update steps on this batch. The learning rate is set to \(1 \times 10^{-6}\) unless otherwise specified.

For baseline methods such as SAPO and GRPO, we experiment with small batch sizes of 32, 64, and 256, and train for multiple epochs until performance begins to degrade. In the distillation pipeline, we typically train for only 1 epoch or less, adjusting the small batch size among 32, 64, corresponding to 32 and 16 stochastic gradient descent update steps, respectively. When selecting models under different hyperparameters, we use the HMMT Feb 25 dataset as the validation set for selection. In the absence of additional notation, subsequent figures and tables report accuracy on HMMT Feb 25.

For online methods, we adopt fully on-policy training. The sampling temperature is selected from the range of 0.6 to 1.0, while other hyperparameters remain consistent with the offline setting.

For algorithm configurations, we adopt common settings. For instance, in GRPO, the clipping thresholds are set to 0.2 and 0.28 by default. In SAPO, the temperature coefficient for negative samples is set to 1.05. For the cross-entropy method, \(\beta\) is tuned between 0.001 and 0.05, and the tuning options also include whether to aggregate log ratios via summation or averaging. For PPO, we follow the recommendations from prior work~\citep{vcppo,yue2025vapo}: \(\lambda=1\) is set during the value network training phase; a dynamic \(\lambda\) is used during the on-policy network training phase, specifically \(\lambda = \frac{1}{0.05 \times \text{length}}\).

During evaluation, the temperature is set to 0.6 (or 1.0 for Qwen3.5 models) and TopP to 0.95, with all other sampling parameters following the default settings of each model. For evaluation, we use the tools provided by Polaris~\citep{Polaris2025} across all benchmarks, except for HMMT Feb 26 and IMO Answer Bench, where we adopt the code tools from Matharena~\citep{balunovic2025matharena}. For experiments on Qwen3.5, which is capable of solving some challenging problems in the benchmarks but sometimes produces answers that may not exactly match the ground truth, we additionally perform LLM-as-judge evaluation using Seed-2.0-Pro as the judge.

\subsection{KL Efficiency}
\begin{figure}[h]
    \centering
    \includegraphics[width=0.85\textwidth]{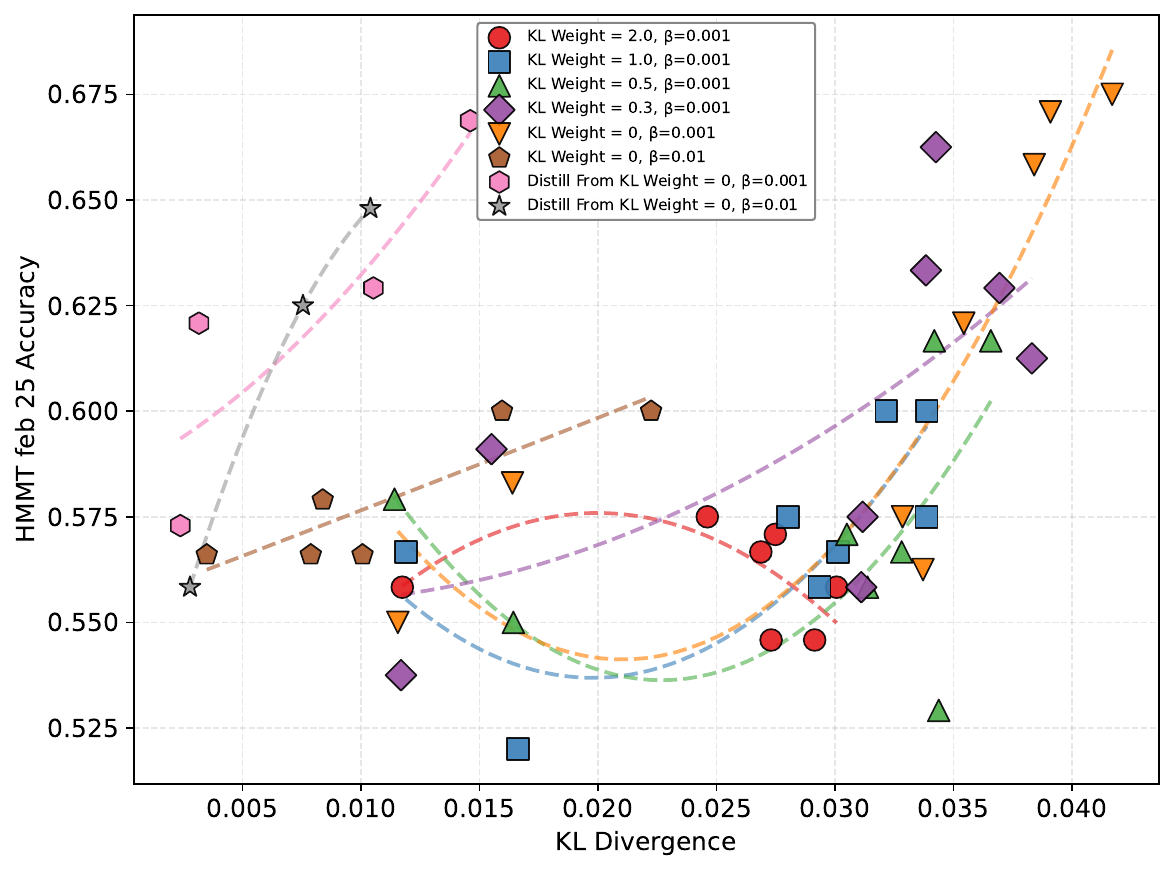}
    \caption{Comparison of KL efficiency before and after distillation on Qwen3-4B.}
    \label{fig:KL}
\end{figure}

We begin by comparing the improvements in KL efficiency brought by distillation. As illustrated in Fig.~\ref{fig:KL}, we first train using cross-entropy (CE) to establish teacher models. 


The figure shows that when the teacher model uses $\beta=0.01$, distillation achieves better performance with low KL consumption, and the student's peak performance exceeds that of the teacher. When $\beta=0.001$, the performance after distillation is comparable to the teacher, but the KL efficiency is higher. Although the proposed surrogate objective only uses the reference model as a support and may not theoretically reduce irrelevant components, experiments demonstrate that it indeed achieves higher KL efficiency and even stronger performance.

The results also show that increasing or decreasing the KL control in the first stage for teacher only appears to affect the upper bound of the teacher model's performance. Regardless of whether the KL control in the first stage is large or small, many irrelevant deviations occur. Model performance only begins to improve significantly after a certain amount of KL divergence has been consumed, specifically in the range from KL=0.01 to KL=0.03. When $\beta=0.001$, performance does not increase notably during this stage despite substantial KL Divergence consumption. This may be due to inherent structural biases in the loss function, which, unlike on-policy gradients, lacks a theoretical guarantee of optimizing towards an improved policy. Furthermore, excessive KL control also limits the optimization effect. For example, in the figure, when the KL weight is 2, the teacher shows no significant improvement.

These observations suggest that it is not easy to simultaneously achieve both high KL efficiency and high sample efficiency in a single-stage optimization. In contrast, by decoupling the process into two stages, our approach achieves better performance on both objectives compared to single-stage training.

\begin{table}[h]
\centering
\caption[Comparison of different policy loss functions on mathematical tasks]{%
    Comparison of training extreme policies with different policy loss functions and subsequent distillation stages on mathematical evaluation sets. AVG@K metrics are reported.%
}\setlength{\tabcolsep}{2pt}
\resizebox{0.8\columnwidth}{!}{%
\begin{tabular}{llcccccc}
\toprule
\multirow{2}{*}{\textbf{Base Model}}
& \multirow{2}{*}{\textbf{Method}}
& \textbf{AIME} & \textbf{AIME} & \textbf{HMMT} & \textbf{Beyond} & \textbf{HMMT}  & \multirow{2}{*}{\textbf{Avg.}} \\
&& \textbf{2024} & \textbf{2025} & \textbf{Feb 25} & \textbf{AIME} & \textbf{Nov 25} \\
\midrule

\multirow{12}{*}{Qwen3-4B-Thinking-2507}
& Base & 84.9 & 81.1 & 56.0 & 53.8 & 66.6 & 68.5  \\
\cmidrule{2-8}
& PPO & 84.8 & 79.8 & 56.5 & 53.5 & 66.0 & 68.1  \\  
\cmidrule{2-8}
& GRPO & 86.8 & 81.2 & 57.8 & 53.5 & 65.7 & 69.0  \\ 
& \quad +Distillation & 86.8 & 80.0 & 61.0 & 54.0 & 64.2 & 69.2  \\
\cmidrule{2-8}
& SAPO & 85.6 & 81.6 & 60.3 & 55.3 & 66.8 & 69.9  \\
& \quad +Distillation & 86.3 & 82.0 & 61.1 & 55.5 & 68.0 & 70.6  \\ 
\cmidrule{2-8}
& CE & 87.3 & 83.6 & 67.6 & 55.7 & 70.4 & 72.9  \\
& \quad +Distillation & 87.9 & 85.1 & 67.1 & 56.3 & 69.9 & \textbf{73.3} \\  
\cmidrule{2-8}
& MSE & 57.2 & 36.6 & 27.7 & 19.5 & 40.8 & 36.4  \\
& \quad +Distillation & 87.5 & 81.9 & 62.7 & 55.3 & 68.0 & 71.1  \\  
\midrule

\multirow{12}{*}{Nanbeige4-3B-2511-Thinking}
& Base & 90.9 & 84.8 & 63.9 & 55.5 & 67.3 & 72.5  \\
\cmidrule{2-8}
& PPO & 91.5 & 86.7 & 65.4 & 55.5 & 70.5 & 73.9  \\
\cmidrule{2-8}
& GRPO & 90.4 & 86.4 & 67.3 & 55.3 & 68.9 & 73.6  \\
& \quad +Distillation & 91.4 & 87.1 & 67.1 & 55.5 & 66.5 & 73.5  \\
\cmidrule{2-8}
& SAPO & 91.8 & 89.1 & 70.2 & 59.0 & 71.1 & 76.2 \\ 
& \quad +Distillation & 91.2 & 89.6 & 73.3 & 61.3 & 72.0 & 77.5  \\
\cmidrule{2-8}
& CE & 91.0 & 88.3 & 70.6 & 61.0 & 72.8 & 76.7  \\
& \quad +Distillation & 90.9 & 88.6 & 71.8 & 63.2 & 73.9 & \textbf{77.7}  \\
\cmidrule{2-8}
& MSE & 63.9 & 48.1 & 31.8 & 15.1 & 26.4 & 37.1  \\
& \quad +Distillation & 91.3 & 87.4 & 68.6 & 59.1 & 68.1 & 74.9  \\
\bottomrule     
\end{tabular}
}
\label{tab:main}
\end{table}

\subsection{Sample Efficiency}
\label{sec:sample_exp}

In Table~\ref{tab:main}, we compare the performance improvements brought by different losses on exactly the same batch of sampled data, and also investigate whether distillation can further improve sample efficiency to achieve performance higher than that of the teacher.

\textbf{The choices of Policy Loss for ERPD.}
First, we analyze the comparison of different loss functions in the off-policy optimization stage of ERPD. Among the considered policy loss functions, SAPO and CE exhibit higher sample efficiency than GRPO. We attribute this to their use of a soft clipping mechanism based on the sigmoid function, which allows each token to receive more complete gradient updates before being clipped.

\textbf{Weak-to-Strong Distillation.}
Second, we discuss the results of weak-to-strong distillation. For policies $\pi_{\theta_e}$ trained with the MSE loss, we observe a significant performance drop compared to the original base model. Nevertheless, when using the log ratio $\log\frac{\pi_{\theta_e}(a \mid s)}{\pi_{\theta_{\text{un}}}(a \mid s)}$ as a token-level reward signal, these degraded policies can still bring substantial improvements to the base model through distillation. This demonstrates a \textbf{weak-to-strong effect}: even policies that perform worse than the student model can provide effective teacher signals when their relative preferences are used for distillation.

\begin{figure}[h]
    \centering
    \begin{subfigure}[b]{0.59\textwidth}
        \centering
        \includegraphics[width=\linewidth]{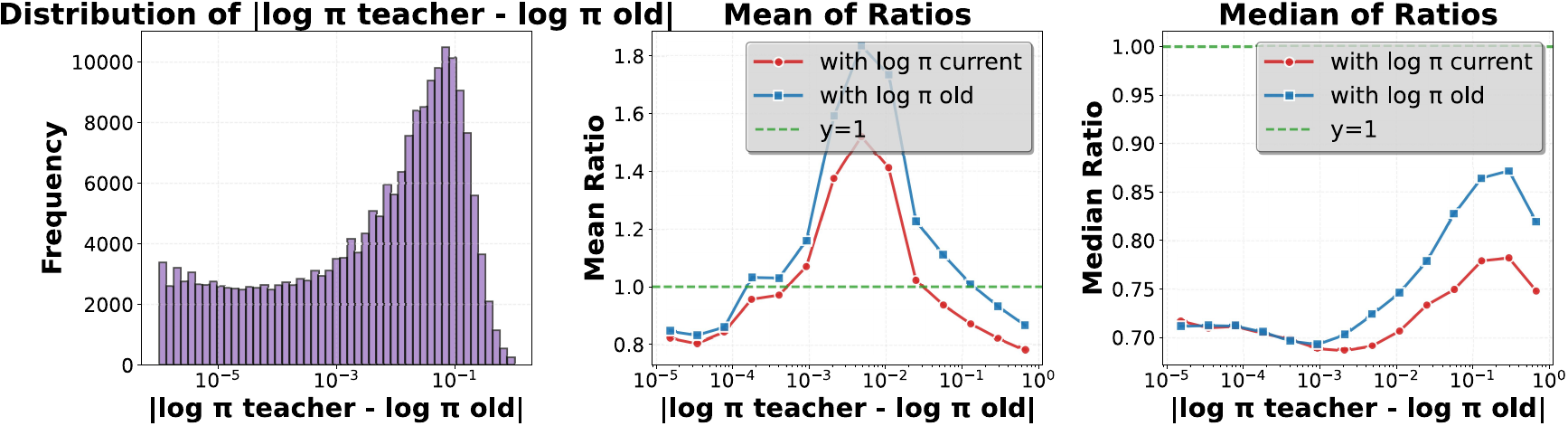}
        \caption{10 distillation steps.}
        \label{fig:on_left}
    \end{subfigure}
    \hfill
    \begin{subfigure}[b]{0.39\textwidth}
        \centering
        \includegraphics[width=\linewidth]{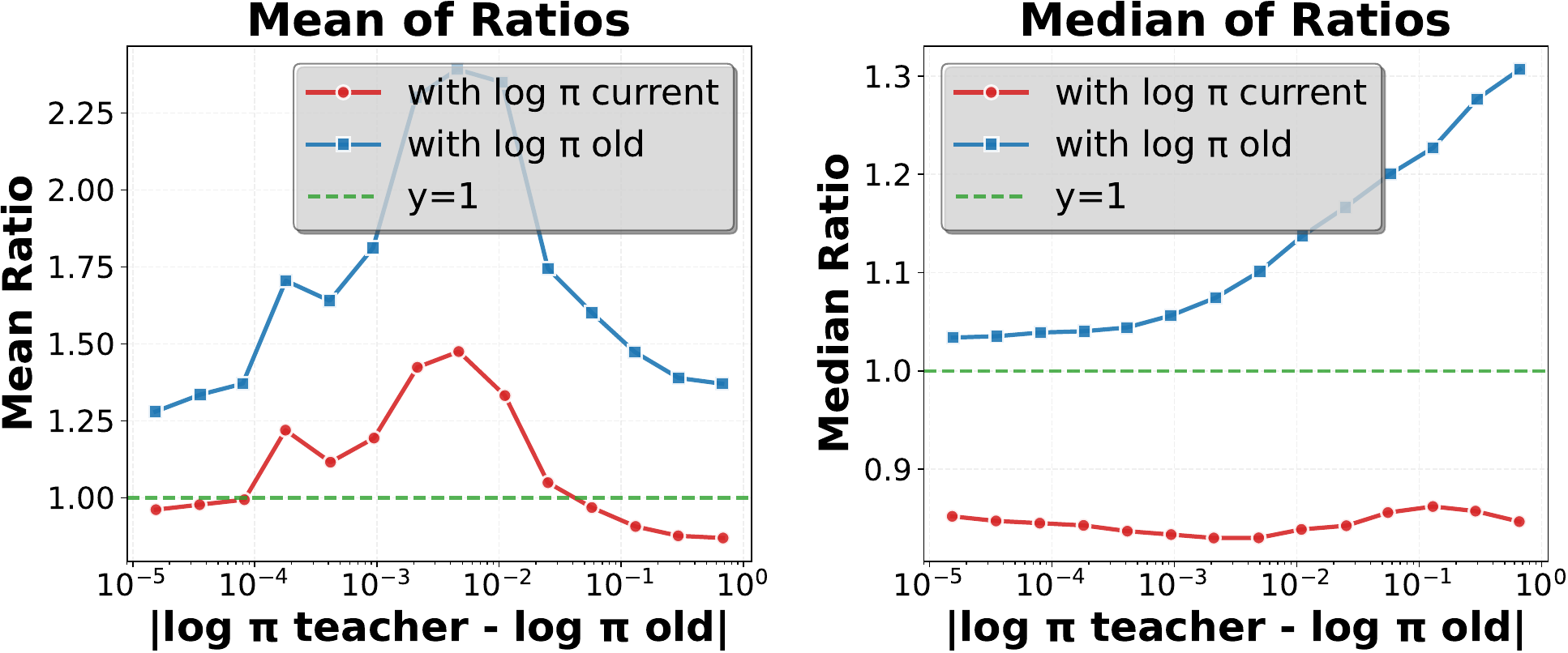}
        \caption{16 distillation steps.}
        \label{fig:on_right}
    \end{subfigure}
    \caption{Comparison of two distillation objectives $\log\frac{\pi_{\theta_e}(a \mid s)}{\pi_{\theta_{\textcolor{myblue}{\text{old}}}}(a \mid s)}$ and $\log\frac{\pi_{\theta_e}}{\pi_{\theta_{}}}$. \textbf{Ratio} in the figure represents \(|\log \pi_{\text{student}} - \log \pi_{\text{old}}| \,/\, |\log \pi_{\text{teacher}} - \log \pi_{\text{old}}|\).
    Teacher setup (Brown line in Fig.~\ref{fig:KL}): Trained on Qwen3-4B using CE ($\beta=0.01$); accuracy: 61.0.
    Using $\log\frac{\pi_{\theta_e}}{\pi_{\theta_{\text{old}}}}$ (\textcolor{myblue}{---}), 
    accuracy reaches \textcolor{myblue}{65.1} (10 steps) and \textcolor{myblue}{63.8} (16 steps). 
    Using $\log\frac{\pi_{\theta_e}}{\pi_{\theta}}$ (\textcolor{myred}{---}), 
    accuracy stays near the teacher at \textcolor{myred}{61.2} and \textcolor{myred}{61.1}. 
    As can be seen from the figure, the blue lines amplifies the components where the teacher deviates significantly, while the red lines tracks the teacher.}
    \label{fig:curvsold}
\end{figure}

\textbf{Surpass Teacher by Distillation.}
Next, we discuss whether distillation can achieve higher performance than the teacher. We observe that in several configurations (e.g., NBG4-3B + CE and NBG4-3B + SAPO), the distilled student even outperforms the teacher, indicating further improved sample efficiency. We attribute this improvement to the use of $\log\frac{\pi_{\theta_e}(a \mid s)}{\pi_{\theta_{\textcolor{myblue}{\text{old}}}}(a \mid s)}$.
From Fig.~\ref{fig:curvsold}, when we modified the log ratio in the distillation signal to $\log\frac{\pi_{\theta_e}(a \mid s)}{\pi_{\theta}(a \mid s)}$, the student converged to performance only comparable to the teacher, without surpassing it. This suggests that $\log\frac{\pi_{\theta_e}(a \mid s)}{\pi_{\theta_{\textcolor{myblue}{\text{old}}}}(a \mid s)}$ may serve as a persistently fixed direction, focusing primarily on the components where the ratio already exhibits large deviations. This components could be beneficial, ultimately enabling the model to exceed the teacher itself.

On the other hand, we suspect that this phenomenon of the student surpassing the teacher may also indicate that the teacher itself could be further improved by tuning its own hyperparameters. In Fig.~\ref{fig:curvsold}, the teacher is trained with CE using $\beta = 0.01$. Using a model that achieved 61 as the teacher yields a distilled student score of 65. However, if we relax $\beta$ to $0.001$ (i.e., a weaker KL constraint), the teacher's performance can reach 67.6. In this case, the student can only reduce ineffective KL and fails to surpass the teacher (pink and orange lines in Fig~\ref{fig:KL}), suggesting that the $\beta = 0.01$ setting is relatively conservative and limits the teacher's ability to fully exploit the samples.


In certain cases (e.g., NBG4-3B with GRPO), distillation yields limited improvement because $\pi_{\theta_e}$ itself achieves only marginal gains over the old model. In such regimes, distillation risks amplifying noise rather than extracting useful structural signals. Empirically, we recommend using sample-efficient objectives such as SAPO or CE to train a strong teacher. However, when these objectives can no longer produce a strong $\pi_{\theta_e}$, weak-teacher strategies such as MSE loss become preferable. We discuss these design choices in the subsequent iterative experiments.

\subsection{Online Experiments}
\label{sec:on}
After comparing the offline KL and sample efficiency, we raise questions about its online/iterative efficiency. 

\begin{table*}[t]
\centering
\caption[Effect of Distillation Strategies]{%
    Comparison of whether to use distillation on Qwen3-4B. In batches 4--6, distillation strategies 2 and 3 from Figure~\ref{fig:1} are adopted, but the teacher model still reports results trained with CE using Strategy 1 for comparison. KL divergence is computed between the current model and the base model.
}
\setlength{\tabcolsep}{1pt}
\resizebox{0.8\linewidth}{!}{%
\begin{tabular}{l c >{\centering\arraybackslash}p{4cm} c c c c c}
\toprule
\multirow{2}{*}{\textbf{Batch}} & \multirow{2}{*}{\textbf{Role}} & \multirow{2}{*}{\textbf{Init. Model}} & \textbf{AIME} & \textbf{HMMT} & \textbf{HMMT} & \multirow{2}{*}{\textbf{Average}} & \multirow{2}{*}{\textbf{KL Div.}} \\
& & & \textbf{2025} & \textbf{Feb 25} & \textbf{Nov 25} \\
\midrule
\multirow{1}{*}{Base}
& - & - & 81.1 & 56.0 & 66.6 & 67.9 & 0 \\
\midrule
\multicolumn{8}{c}{\textit{With Distillation}} \\
\midrule
\multirow{2}{*}{Batch 1}
& Teacher & Base & 83.6 & 67.6 & 70.4 & 73.8 & 0.041 \\
& Student & Base & 82.5 & 62.9 & 68.1 & 71.1 & 0.010 \\
\midrule
\multirow{2}{*}{Batch 2}
& Teacher & Batch 1 Student & 85.8 & 68.7 & 72.5 & 75.6 & 0.044 \\
& Student & Batch 1 Student & 84.5 & 67.9 & 68.5 & 73.6 & 0.026 \\
\midrule
\multirow{2}{*}{Batch 3}
& Teacher & Batch 2 Student & 86.3 & 70.3 & 75.1 & 77.2 & 0.042 \\
& Student & Batch 2 Student & 84.5 & 71.2 & 73.5 & 76.4 & 0.037 \\
\midrule
\multicolumn{8}{c}{\textit{Using Strategy 3, Selecting Batch 1 Student}} \\
\multirow{2}{*}{Batch 4}
& Teacher & Batch 3 Student & 86.1 & 71.4 & 75.2 & 77.5 & 0.043 \\
& Student & Batch 3 Student & 87.9 & 71.7 & 78.3 & 79.3 & 0.062 \\
\midrule
\multicolumn{8}{c}{\textit{Using Strategy 2: MSE}} \\
\multirow{2}{*}{Batch 5}
& Teacher & Batch 4 Student & 89.3 & 72.0 & 77.0 & 79.4 & 0.066 \\
& Student & Batch 4 Student & 88.0 & 72.8 & 79.0 & 79.9 & 0.065 \\
\midrule
\multicolumn{8}{c}{\textit{Using Strategy 2: MSE}} \\
\multirow{2}{*}{Batch 6}
& Teacher & Batch 5 Student & 90.4 & 72.2 & 77.9 & 80.2 & 0.077 \\
& Student & Batch 5 Student & 89.2 & 73.6 & 79.0 & 80.6 & 0.073\\
\midrule
\multicolumn{8}{c}{\textit{Without Distillation}} \\
\midrule
\multirow{1}{*}{Batch 1 (a)}
&  & Base & 83.6 & 67.6 & 70.4 & 73.8 & 0.041 \\
\multirow{1}{*}{Batch 2 (a)}
&  & Batch 1 (a) & 84.1 & 67.9 & 70.0 & 74.0 & 0.054 \\
\midrule
\multirow{1}{*}{Batch 1 (b)}
&  & Base & 83.3 & 61.2 & 67.5 & 70.6 & 0.022 \\
\multirow{1}{*}{Batch 2 (b)}
&  & Batch 1 (b) & 79.2 & 61.2 & 65.0 & 68.4 & 0.030 \\
\bottomrule
\end{tabular}
}
\label{tab:batch_inheritance}
\end{table*}

\textbf{Iterative Distillation brings better asymptotic performance.}
In the iterative experiments, compared to the settings in Tab.~\ref{tab:main}, where 16 or 32 distillation steps were used, we recommend adopting a more conservative approach with 10 distillation steps. Our experiments show that this leads to better asymptotic performance.

As shown in Table~\ref{tab:batch_inheritance}, in Batch 1, because the number of distillation steps was chosen very conservatively (10 steps), the student did not fully fit the teacher. The average score was 71.1, worse than the teacher's 73.8, but the KL divergence cost was only one-quarter of that of the teacher. 
From Batch 1 to 3, model performance shows a steady upward trend across batches. Eventually, on the student model of Batch 3, the average score reaches 76.4, compared to the base score of 67.9.
In contrast, in the control experiment where distillation was not used to reduce KL divergence cost, we selected two checkpoints of the teacher model to continue training, with KL divergence costs of 0.022 and 0.041, respectively. The experiments show that performance either improved only marginally (e.g., setting (a) increased only from 73.8 to 74.0) or even degraded (e.g., setting (b) dropped from 70.6 to 68.4). This suggests that relying solely on iterative offline training without distillation makes it difficult to steadily improve performance and may even lead to error accumulation.

\textbf{If a strong teacher cannot be directly obtained, one resorts to Strategies 2 and 3.} Next, we discuss recommendations for constructing the teacher signal.
At the beginning of Batch 4, the average score of the Batch 4 teacher (77.5) is very close to that of the Batch 3 teacher (77.2). Preliminary experiments showed that continuing to distill with the Batch 4 teacher could not further improve performance. Therefore, we switched to Strategies 2 and 3 illustrated in Figure~\ref{fig:1}. For Batch 4, Strategy 3 used the signal $\log\frac{\pi_{\theta_{\text{Batch 4 teacher}}}(a \mid s)}{\pi_{\theta_{\text{Batch 1 student}}}(a \mid s)}$. It was observed that after applying this strategy, the KL divergence cost became high. This is because optimizing $\log\frac{\pi_{\theta_{\text{Batch4 teacher}}}(a \mid s)}{\pi_{\theta_{\text{Batch 1 student}}}(a \mid s)}$ increases the distance from $\pi_{\theta_{\text{Batch 1 student}}}(a \mid s)$. The experiments also revealed that after this round of optimization, the generation length of the model increased significantly. This may be due to the differential signal in the log-ratio: training pushes the model away from a weaker policy with shorter generation length, which is a different philosophy from Strategy 1 — shifting from learning toward a strong teacher to moving away from a weak teacher.

For Batches 5 and 6, Strategy 2 use the  MSE loss signal introduced in the methodology section, i.e., $\log\frac{\pi_{\theta_e}(a \mid s)}{\pi_{\theta_{\text{un}}}(a \mid s)}$. The results show that when the CE teacher achieves only a small improvement,  using a weak teacher for distillation leads to better results. Strategy 2 may work through different intrinsic mechanisms compared to Strategy 1. This will be further discussed in the next section, which presents experimental results on signal ensemble.

\begin{figure}[h]
    \centering
    \includegraphics[width=0.9\textwidth]{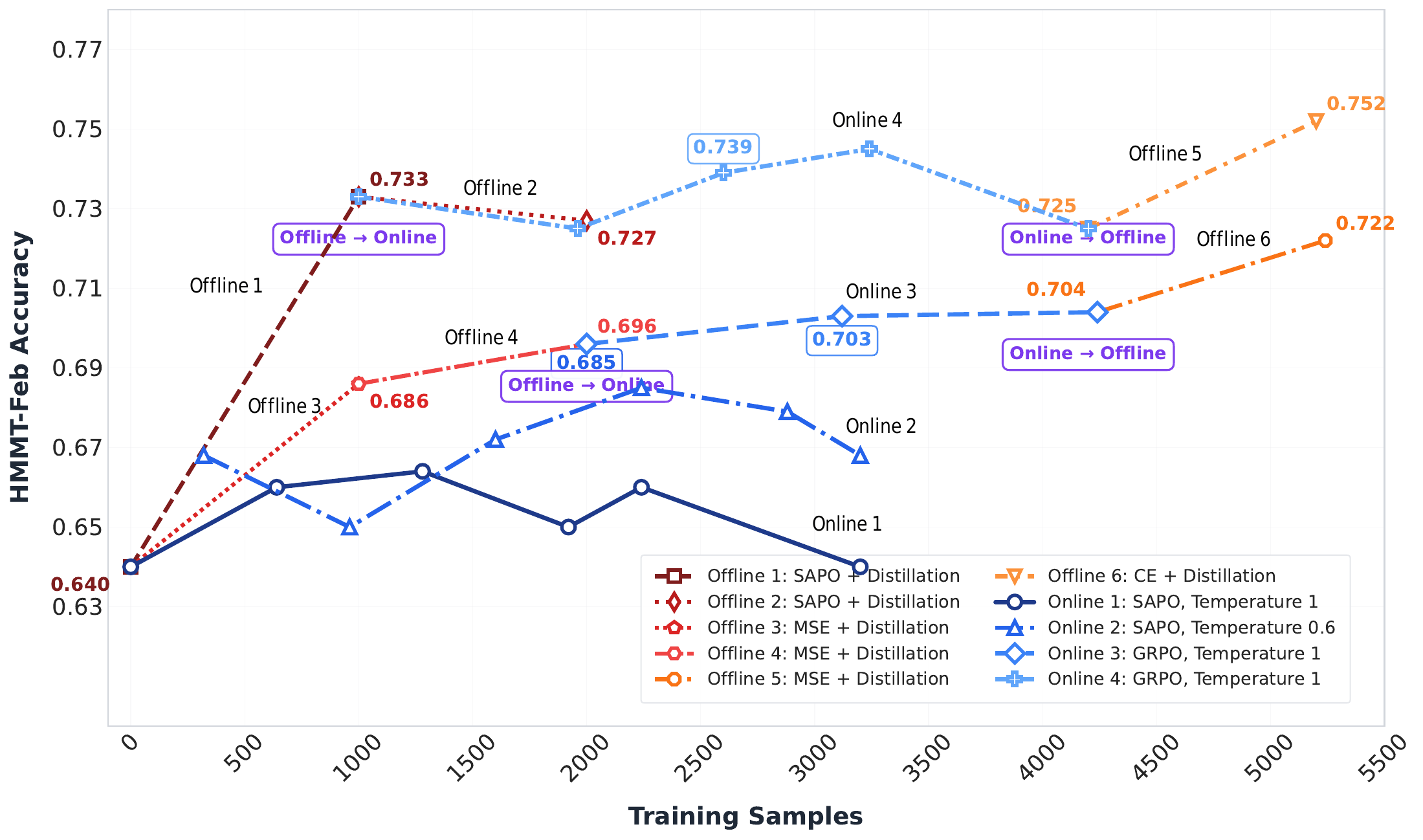}
    \caption{Switching between online and offline stage bring better performance on NBG4-3B.}
    \label{fig:on}
\end{figure}

\textbf{Online-offline Switching.}
Aside from signal construction, when the model can no longer be improved through offline methods, trying to switch between offline and online may be a viable option:
As shown in Fig.~\ref{fig:on}, the online phases (Online-1 and Online-2) show only slow improvement relative to offline methods on NBG4-3B. On Qwen3-4B, on-policy can even fail (Tab.~\ref{tab:starting_point}).
Offline stages, for example Offline-1, in contrast, show a clear improvement over Online-1. However, when Offline-2 is performed by recollecting another 1K samples using this strong model (score 73.3), no additional improvement is observed. In the other offline phases (Offline-3 and Offline-4), performance improves more consistently, but the gains are smaller than those achieved in the initial offline stage.

After inserting an intermediate online phase, subsequent offline stages (Offline-5 and Offline-6) again demonstrate improved sample efficiency.
Overall, these results suggest that switching between offline and online training can be beneficial for achieving sustained performance improvement.

\begin{table}[t]
\centering
\caption{Naive on-policy GRPO performs poorly on Qwen3-4B.}
\label{tab:starting_point}
\small

\begin{tabular}{@{}lccc@{}}
\toprule
Init. model & \textbf{Step 0} & \textbf{Step 25} & \textbf{Step 50} \\
\midrule
Base model & 56.0 & 57.2 & 53.2 \\
\bottomrule
\end{tabular}
\qquad
\begin{tabular}{@{}lccc@{}}
\toprule
Init. model & \textbf{Step 0} & \textbf{Step 40} & \textbf{Step 60} \\
\midrule
Batch 4 student & 71.6 & 70.4 & 71.4 \\
\bottomrule
\end{tabular}

\end{table}

\textbf{Optimization Time Comparison.}
We also compare the optimization time efficiency with online methods in Tab.~\ref{tab:stage_comparison}. Offline optimization requires repeatedly using the data, with the main time consumption lying in teacher training, which needs multiple epochs to converge. In contrast, the distillation stage requires very few steps and consumes little time. In large-scale model training, compared to online methods, offline optimization may offer certain advantages in terms of framework—for example, the training stage does not require loading a generative model, which can save some GPU memory overhead. During the distillation stage, the teacher's signals can also be obtained through preprocessing, avoiding out-of-memory issues.

\begin{table}[h]
\centering
\caption{ 
Efficiency Comparison across Optimization Steps and Time on NBG4-3B. On-policy and ours denote Online-2 and Offline-1 in Fig.~\ref{fig:on}, respectively.}
\label{tab:stage_comparison}
\small
\begin{tabular}{@{}llcc@{}}
\toprule
\textbf{Stage} & \textbf{Step} & \textbf{Score} & \textbf{Cumulative time} \\
\midrule
\multirow{4}{*}{\parbox{4cm}{\raggedright On-policy \\[-2pt] \scriptsize (800\,s rollout + 300\,s update per step)}} 
  & 10  & 66.8 & $\sim$3\,h \\
  & 50  & 67.2 & $\sim$15\,h \\
  & 70  & 68.5 & $\sim$21\,h \\
  & 100 & 66.8 & $\sim$31\,h \\
\midrule
\multirow{3}{*}{\parbox{4cm}{\raggedright Ours, first stage \\[-2pt] \scriptsize (300\,s per step + offline rollout 7h)}} 
  & 40  & 68.9 & $\sim$ 7+ 3\,h \\
  & 80  & 68.3 & $\sim$ 7 + 7\,h \\
  & 100 & 70.5 & $\sim$ 7 + 8\,h \\
\midrule
\multirow{3}{*}{\parbox{4cm}{\raggedright Ours, distill \\[-2pt] \scriptsize (400\,s per step +  the first stage 17h)}} 
  & 10  & 67.2 & $\sim$17 + 1\,h \\
  & 16  & 68.3 & $\sim$17 + 2\,h \\
  & 32  & \textbf{73.3} & $\sim$17 + 3\,h \\
\bottomrule
\end{tabular}
\end{table}

\textbf{KL and Entropy Controls during Distillation.}
In Tab.~\ref{tab:KL散度}, we compare different KL control strategies to assess their impact on KL efficiency and effectiveness, covering two categories: the number of optimization steps and explicit KL losses.  
As shown, for different numbers of steps, the model with 16 steps does not achieve complete fitting, while the model with 32 steps achieves a better performance but also more KL.  The KL consumption can be mitigated by introducing KL loss. Appropriate KL loss weight, such as 0.05 or 0.1, can also achieve a relatively good balance. However, it is worth noting that if the goal is to minimize KL divergence, selecting the 16-step result or using a higher KL loss value of 0.2 is also reasonable, as it shows some improvement over the base model while consuming very little KL.

\begin{table}[h]
\centering
\caption{Comparison of Different KL Divergence Control Methods on NBG4-3B}
\label{tab:KL散度}
\resizebox{0.55\linewidth}{!}{%
\begin{tabular}{lcccc}
\toprule
\multirow{2}{*}{\textbf{Method}}
& \multirow{2}{*}{\textbf{Steps}} 
& \textbf{KL Loss} 
& \textbf{HMMT} 
& \multirow{2}{*}{\textbf{KL Div.}}  \\
& & \textbf{Weight}  &   \textbf{Feb 25}   &  \\
\midrule

Base                    & --   & --    & 63.9 & 0.000 \\
\midrule
SAPO                    & 60   & --    & 68.1 & 0.016 \\
SAPO                    & 120  & --    & 70.2 & 0.023 \\
\midrule
SAPO + Distillation          & 16   & --    & 68.3 & \textbf{0.009} \\
SAPO + Distillation          & 20   & --    & 70.2 & 0.017 \\
SAPO + Distillation          & 32   & --    & \textbf{73.3} & 0.026 \\

\addlinespace[0.2em] 
SAPO + Distillation          & 32   & 0.2   & 70.2 & 0.014 \\
SAPO + Distillation          & 32   & 0.05   & 69.6 & 0.020  \\
SAPO + Distillation          & 32   & 0.001   & 71.7 & 0.018 \\
\bottomrule     
\end{tabular}
}
\end{table}

\begin{table*}[h]
\centering
\caption[Effect of Distillation Strategies]{In contrast to KL loss, which constrains distillation effectiveness, Entropy loss does not hurt distillation performance. In fact, when a relatively large entropy loss is applied during distillation, entropy increases, and performance can also rise accordingly.}
\begin{subtable}[t]{0.48\linewidth}
\centering
\caption{w/o Entropy Loss}
\setlength{\tabcolsep}{3pt}
\resizebox{\linewidth}{!}{%
\begin{tabular}{l c c c c}
\toprule
\textbf{Batch} & \textbf{AIME 2025} & \textbf{HMMT Nov 25} & \textbf{Entropy} & \textbf{KL Div.} \\
\midrule
Base & 81.1 & 66.6 & 0.20 & 0 \\
\midrule
Batch 1 & 82.5 & 68.1 & \textbf{0.19} & 0.010 \\
Batch 2 & 84.5 & 68.5 & \textbf{0.16} & 0.026 \\
Batch 3 & 84.5 & 73.5 & \textbf{0.14} & 0.037 \\
\bottomrule
\end{tabular}%
}
\end{subtable}
\hfill
\begin{subtable}[t]{0.48\linewidth}
\centering
\caption{w/ Entropy Loss}
\setlength{\tabcolsep}{3pt}
\resizebox{\linewidth}{!}{%
\begin{tabular}{l c c c c}
\toprule
\textbf{Batch} & \textbf{AIME 2025} & \textbf{HMMT Nov 25} & \textbf{Entropy} & \textbf{KL Div.} \\
\midrule
Base & 81.1 & 66.6 & 0.20 & 0 \\
\midrule
Batch 1 & 83.7 & 71.0 & \textbf{0.19} & 0.014 \\
Batch 2 & 88.3 & 74.5 & \textbf{0.21} & 0.023 \\
Batch 3 & 85.6 & 76.0 & \textbf{0.25} & 0.032 \\
\bottomrule
\end{tabular}%
}
\end{subtable}
\label{tab:ent}
\end{table*}

For entropy control, in our iterative experiments (Tab.~\ref{tab:ent}), we continue to use standard CE training for the teacher, but apply a larger entropy loss with a weight of 0.5 during each distillation step. Unlike the typical pattern where stronger performance is accompanied by increasing KL divergence from the reference model, the performance gains do not rely on entropy reduction. As shown in the table, even with sustained or increased entropy, the model can still distill effective improvement signals, outperforming the counterpart without entropy loss. 
This also suggests that some components driving entropy reduction may be irrelevant to model improvement; more effective methods for filtering out such impurities could further enhance distillation efficacy and asymptotic performance.

\subsection{Unlearned Extreme Region Policy}
Based on the MSE teacher, surprisingly effective weak-to-strong generalization is achieved. We will conduct detailed experiments on signal construction ablation, ensemble effects, loss ablation, and generalizability in what follows.

\label{sec:expmse}

\textbf{Ablation on Signal Construction.}
Table~\ref{tab:ablation_signal} presents the ablation study on distillation signal construction. We systematically vary the training steps of the teacher model (numerator) and the unlearning model (denominator) to investigate their individual contributions.

During development, we generally choose a model with as many training steps as possible as the teacher, without performing very careful hyperparameter tuning. The factor that has a greater impact is the choice of the unlearned policy. As shown in the table, if the unlearned policy is not chosen properly, the performance will degrade. However, experiments also show that performing more hyperparameter tuning on the teacher model can achieve the best results, although it is relatively less sensitive compared to the choice of the unlearned policy.
The optimal configuration emerges when the teacher is trained for 50 steps and the unlearned policy for 10 steps, yielding the highest accuracy of 64.3.

Besides, replacing the teacher with the current starting model (``Curr.'' in numerator) or the unlearned policy with the current model (``Curr.'' in denominator) both lead to suboptimal performance (59.1 and 61.3, respectively), indicating that the explicit separation between teacher and unlearn trajectories is crucial. Notably, omitting either term entirely (``None'') results in significant degradation, with the unlearn-only variant dropping to 56.3, nearly equivalent to the base model. These results collectively demonstrate that both components of the ratio contribute synergistically to the distillation efficacy, and their relative optimization levels require careful balancing.

\begin{table}[h]
\centering
\small
\setlength{\tabcolsep}{3.5pt}
\caption{Ablation on signal construction on Qwen3-4B. 
         The distillation signal is $\log \frac{\pi_{\theta_e}(a \mid s)}{\pi_{\theta_{\text{un}}}(a \mid s)}$. 
         We vary the training steps of the teacher (numerator) and the unlearned policy (denominator). 
         ``Curr.'' substitutes the current starting model $\pi_{\theta_{\text{old}}}$ at that position; 
         ``None'' omits the corresponding term (yielding $\log \pi_{\theta_e}$ or $-\log \pi_{\theta_{\text{un}}}$ alone); 
         ``--'' denotes the base model.}
\label{tab:ablation_signal}
\begin{tabular}{@{}lccccccccccc@{}}
\toprule
& \multicolumn{11}{c}{Training steps of teacher and unlearning models} \\
\cmidrule(lr){2-12}
Teacher  & -- & 70 & 70 & 70 & 60 & 50 & 40 & Curr. & 10 & 10 & None \\
Unlearn  & -- & 10 & 20 & 30 & 10 & 10 & 10 & 50 & Curr. & None & 50 \\
\midrule
HMMT Feb 25 & 56.0 & 60.1 & 59.1 & 53.1 & 61.0 & \textbf{64.3} & 59.5 & 59.1 & 61.3 & 56.3 & 59.3 \\
\bottomrule
\end{tabular}
\end{table}

\begin{table*}[h]
\caption[Impact of different teacher reward designs on Qwen3-4B]{%
    Impact of different teacher reward designs on the Qwen3-4B model. A checkmark indicates the use of a teacher model fine-tuned from the corresponding \textbf{Unlearned Policy} or \textbf{Old Policy}. \textbf{Recovery Loss} refers to the loss employed to recover the model from the unlearned policy, while \textbf{Old Policy Loss} refers to the loss used to derive a strong teacher from the old policy. $\beta$ controls the strength of the KL divergence constraint in the CE loss, where ${\beta_o}$ is for the old policy and ${\beta_u}$ is for the unlearned policy. Reported results are the performance of the student model after distillation.%
}\centering
\small
\setlength{\tabcolsep}{4pt}
\resizebox{0.95\columnwidth}{!}{
\begin{tabular}{cccccccccc}
\toprule
\multirow{2}{*}{\textbf{Unlearned Policy}} 
& \multirow{2}{*}{\textbf{Old Policy}}
& \multirow{2}{*}{\textbf{Recovery Loss }}
& \multirow{2}{*}{$\boldsymbol{\beta_f}$}
& \multirow{2}{*}{\textbf{Old Policy Loss}}
& \multirow{2}{*}{$\boldsymbol{\beta_o}$}
& \textbf{AIME} & \textbf{AIME} & \textbf{HMMT} & \textbf{Beyond} \\
& & & & & & \textbf{2024} & \textbf{2025} & \textbf{Feb 25} & \textbf{AIME} \\
\midrule
-- & -- & -- & -- & -- & -- & 84.9 & 81.1 & 56.0 & 53.8 \\
\midrule
\checkmark & -- & MSE & -- & -- & -- 
& 87.5 & 81.9 & 62.7 & 55.3 \\
-- & \checkmark & -- & -- & CE & $5\!\times\!10^{-2}$ &
89.2 & 84.1 & 63.8 & 54.4 \\
-- & \checkmark & -- & -- & CE & $10^{-3}$ &
87.9 & 85.1 & 67.1 & 56.3 \\
\midrule
\checkmark & \checkmark & MSE & -- & CE & $5\!\times\!10^{-2}$ 
& \underline{89.3} & \underline{85.2} & 66.7 & \underline{56.4} \\
\midrule
\checkmark & \checkmark & CE & $10^{-3}$ & CE & $10^{-3}$ 
& 87.5 & 83.5 & 66.5 & 55.9 \\
\checkmark & \checkmark & CE & $10^{-2}$ & CE & $5\!\times\!10^{-2}$ 
& 89.1 & 84.3 & \underline{66.8} & 55.1 \\
\checkmark & \checkmark & CE & $10^{-3}$ & CE & $5\!\times\!10^{-2}$ 
& \textbf{90.1} & \textbf{85.4} & \textbf{69.0} & \textbf{56.7} \\
\bottomrule
\end{tabular}
}
\label{tab:esm}
\end{table*}
\textbf{Teacher Ensembles.}
As shown in Table~\ref{tab:esm}, we construct ensemble signals from distinct teachers. \textbf{Old Policy} denotes distillation via strategy 1 in Figure~\ref{fig:1}, where the teacher is a policy trained with the policy loss (already improved over the base model). \textbf{Unlearned Policy} denotes distillation via the strategy 2 in Figure~\ref{fig:1}, where the teacher is trained with MSE loss and has undergone a unlearning phase, making it weaker than the base model.

For the concrete implementation of the ensemble method, we first train several steps with Teacher 1's reward, then several steps with Teacher 2's reward, selecting the step count (16 or 32) based on validation performance.
When both teacher signals are provided simultaneously, as in the last four rows of the table, the resulting (MSE+CE) ensemble outperforms using either signal alone. This indicates that weak and strong teachers can be effectively combined to improve sample efficiency, and their knowledge exhibits complementarity.


In the reward signal constructed from the policy trained with MSE loss, the unlearned policy $\pi_{\theta_{\text{un}}}$ plays the role of $\pi_{\theta_{\text{old}}}$ in the log-ratio.
An open research question is \textbf{whether MSE loss is strictly necessary for training the teacher model in Strategy 2}. From this perspective, we retain a unlearned policy as the starting model but train it with CE rather than MSE.
Comparing the last row with the fourth-to-last row, we observe that continuing to train the unlearned policy with CE yields even better ensemble performance than using the original MSE loss. This suggests that MSE may not be crucial; what matters is starting from a weaker model, as its larger room for improvement may provide more pronounced aggressive improvement signals.

Furthermore, the ensemble combining strong KL divergence constraints on the old policy with weak KL divergence constraints on the unlearned policy produces the best ensemble result, namely the last row of Table~\ref{tab:esm}. Notably, this ensemble delivers a remarkable improvement of up to 13 percentage points on HMMT Feb 25. In the table, $\beta$ denotes the strength of the KL divergence constraint: a larger $\beta$ implies stronger regularization toward the reference model. Comparing the last, third-to-last, fifth-to-last, and sixth-to-last rows, we see that although the teacher in the fifth-to-last row is on average stronger than that in the sixth-to-last row, its effect when ensembled with the unlearned policy (third-to-last row) is nevertheless weaker than the last row.
This suggests that different teachers may require updates of varying conservatism to provide complementary knowledge for effective ensemble. We leave the systematic exploration of multi-stage or multi-teacher ensembles to future work.

\begin{table}[h]
\centering
\small
\caption{Recovering unlearned policy using CE and SFT as the teachers.}
\label{tab:qwen_ablation_split}
\setlength{\tabcolsep}{6pt}
\begin{tabular}{@{}llcc@{}}
\toprule
\textbf{Unlearned Policy} & \textbf{Recovery Loss} & \textbf{HMMT Feb 25} & \textbf{HMMT Nov 25} \\
\midrule
\multirow{2}{*}{lr=$2\mathrm{e}{-6}$, $w=0.5$} & SFT & 57.6 & 66.4 \\
                               & CE  & \textbf{62.5} & \textbf{71.0} \\
\cmidrule(lr){2-4}
\multirow{2}{*}{lr=$2\mathrm{e}{-6}$, $w=1$}   & SFT & 53.5 & 64.9 \\
                               & CE  & \textbf{64.1} & 70.4 \\
\cmidrule(lr){2-4}
\multirow{2}{*}{lr=$5\mathrm{e}{-6}$, $w=1$}   & SFT & \textbf{63.5} & 67.9 \\
                               & CE  & 62.6 & 67.9 \\
\bottomrule
\end{tabular}
\end{table}

    \begin{figure}[h]
    \centering
    \begin{subfigure}[b]{0.32\linewidth}
        \centering
        \includegraphics[width=\linewidth]{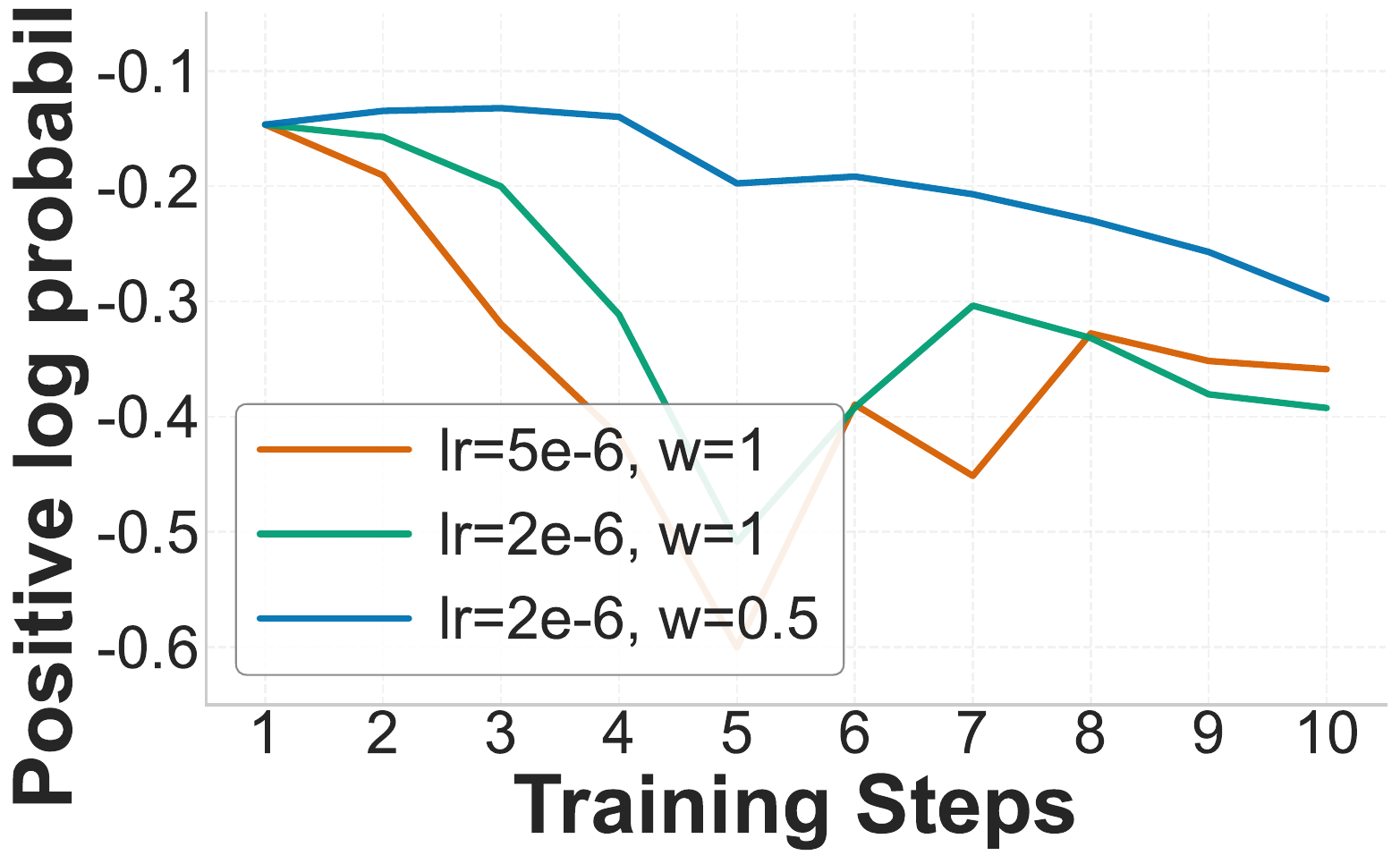}
        \caption{MSE training is first used to obtain an unlearned policy.}
        \label{fig:sfta}
    \end{subfigure}
    \hfill
    \begin{subfigure}[b]{0.62\linewidth}
        \centering
        \includegraphics[width=\linewidth]{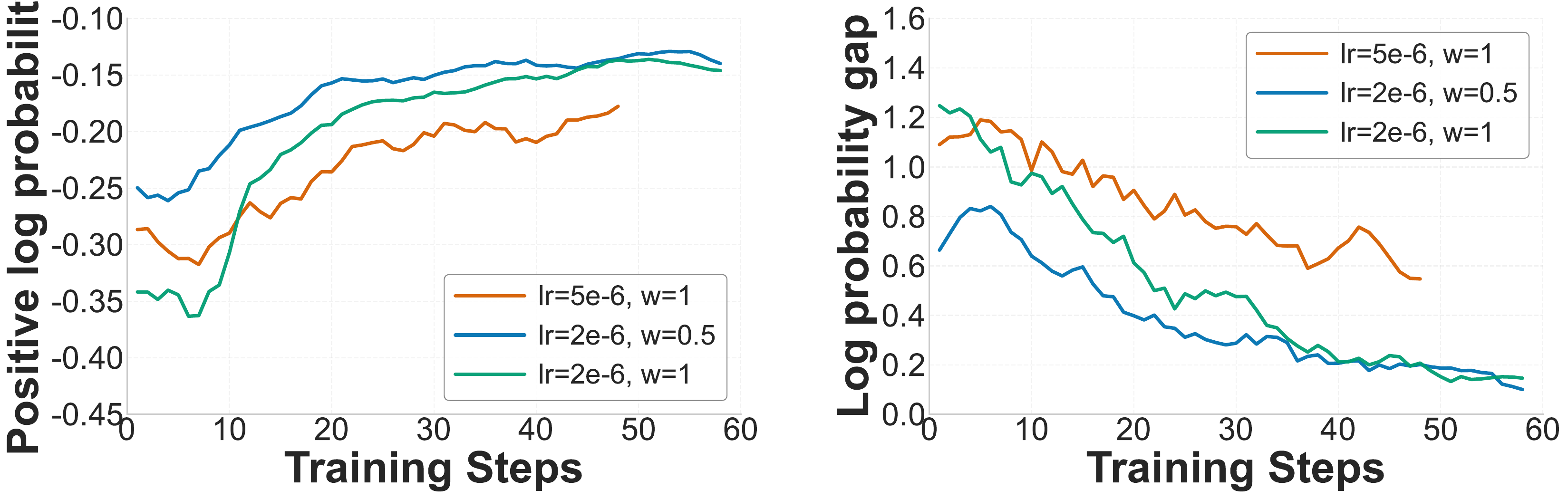}
        \caption{Performing SFT on previous 10-step unlearned policy.}
        \label{fig:sftb}
    \end{subfigure}
    \caption{An ablation experiment where the loss in the recovery stage is replaced from MSE to SFT. A larger learning rate during the unlearning can lead to better positive-negative separation during the SFT stage.}
    \label{fig:sft}
\end{figure}

\textbf{Is SFT on positive examples from the unlearned model 
sufficient for weak-to-strong recovery?} We conduct an ablation study for this question, as shown in Tab.~\ref{tab:qwen_ablation_split}. Here, a larger learning rate indicates stronger unlearning, and $w$ denotes the weight assigned to negative examples relative to positive examples---a larger $w$ leads to stronger unlearning. We find that SFT is effective only when the model has undergone severe unlearning. The reason can be identified from the Fig.~\ref{fig:sft}: Fig.~\ref{fig:sfta} shows that the orange curve experiences the strongest unlearning phase, while Fig.~\ref{fig:sftb} reveals that with weaker unlearning, although only positive examples are fine-tuned during SFT, the gap between negative and positive examples gradually diminishes, bringing them to nearly the same level. In contrast, a more severely unlearned model can appropriately preserve the gap between positive and negative examples, achieving performance comparable to CE in the table.

\textbf{Broad Efficacy of MSE across Models, Tasks, and Iterations.}
In addition to the two primary experimental models, Tab.~\ref{tab:other} reports results on three additional models. The table shows that for three models with longer generation lengths, the method yields benefits. However, for models like Ace-7B and JustRL-1.5B, which have shorter generation lengths, the improvement is relatively small. It is hypothesized that generation length may influence the effectiveness of the method to some extent.

\begin{table}[h]
\caption{Additional model results using the MSE training strategy}
\centering
\small
\setlength{\tabcolsep}{0.3pt}
\begin{tabular}{lccccccc}
\toprule
\multirow{2}{*}{\textbf{Base Model}}
& \multirow{2}{*}{\textbf{Method}}
& \textbf{AIME} & \textbf{AIME} & \textbf{HMMT} & \textbf{AIME24}  \\
&& \textbf{2024} & \textbf{2025} & \textbf{Feb 25} & \textbf{Avg. Length}  \\
\midrule
\multirow{2}{*}{Llama-3.1-Nemotron-Nano-4B-v1.1}
& Base & 71.0 & 54.3 & 40.8 & 19.8K  \\
& MSE+Distillation & 71.8 & 58.6 & 44.3 & 18.0K   \\
\midrule
\multirow{2}{*}{AceReason-Nemotron-1.1-7B}
& Base & 73.3 & 65.0 & 45.0 & 14.8K  \\
& MSE+Distillation & 73.9 & 66.2 & 44.8 & 14.9K   \\
\midrule
\multirow{2}{*}{JustRL-Nemotron-1.5B}
& Base & 69.2 & 60.8 & 39.3 & 12.8K  \\
& MSE+Distillation & 69.7 & 61.8 & 40.1 & 12.9K  \\
\midrule
\multirow{2}{*}{Qwen3-4B}
& Base & 84.9 & 81.1 & 56.0 & 19.8K  \\
& MSE+Distillation & 87.5 & 81.9 & 62.7 & 20.8K  \\
\midrule
\multirow{2}{*}{NBG4-3B}
& Base & 90.9 & 84.8 & 63.9 & 24.8K  \\
& MSE+Distillation & 91.2 & 86.6 & 67.6 & 23.3K  \\
\bottomrule     
\end{tabular}
\label{tab:other}
\end{table}

\begin{table}[h]
\centering
\small
\setlength{\tabcolsep}{2pt}
\caption{Performance comparison on code and math tasks on Qwen3.5 models. For the code task , numbers in parentheses indicate the learning rate of the unlearned policies. For math tasks, results show iterative experiments.}
\label{tab:unified_code_math}
\begin{tabular}{@{}llcccccc@{}}
\toprule
& & \multicolumn{1}{c}{\textbf{Code}} & \multicolumn{5}{c}{\textbf{Math}} \\
\cmidrule(lr){3-3} \cmidrule(lr){4-8}
\multirow{2}{*}{\textbf{Model}} & \multirow{2}{*}{\textbf{Config}} & \multirow{2}{*}{\textbf{LiveCodeBenchV6}} & \multicolumn{3}{c}{\textbf{HMMT}} & \textbf{IMO} & \multirow{2}{*}{\textbf{Avg}} \\
\cline{4-6}
& & & \textbf{Nov 25} & \textbf{Feb 25} & \textbf{Feb 26} & \textbf{Answer Bench} & \\
\midrule
\multicolumn{8}{@{}l}{\textit{Qwen3.5-9B}} \\
\midrule
Base & -- & 56.1 & 84.0 & 83.5 & 72.7 & -- & 80.1 \\
CE & -- & 50.7 & 83.3 & 84.1 & 75.0 & -- & 80.8 \\
\midrule
\multirow{6}{*}{MSE+Distillation } 
 & LR=1e-6 & 40.2 & -- & -- & -- & -- & -- \\
 & LR=5e-6 & 63.7 & -- & -- & -- & -- & -- \\
 & LR=1e-5 & 62.4 & -- & -- & -- & -- & -- \\
\cmidrule{2-8}
 & Batch 1 & -- & 84.1 & 85.6 & 78.8 & -- & 82.8 \\
 & Batch 2 & -- & 88.3 & 88.7 & 82.2 & -- & 86.4 \\
 & Batch 3 & -- & 87.3 & 88.9 & 83.8 & -- & 86.7 \\
\midrule
\multicolumn{8}{@{}l}{\textit{Qwen3.5-27B}} \\
\midrule
Base & -- & -- & 89.8 & 92.0 & 84.3 & 79.9 & 86.5 \\
\midrule
\multirow{3}{*}{MSE+Distillation } 
 & Batch 1 & -- & 93.3 & 94.1 & 83.5 & 84.0 & 88.7 \\
 & Batch 2 & -- & 92.2 & 93.0 & 88.0 & 85.2 & 89.5 \\
 & Batch 3 & -- & 94.1 & 94.1 & 89.2 & 86.3 & 90.9 \\
\bottomrule
\end{tabular}
\end{table}

Additionally, Tab.~\ref{tab:unified_code_math} reports results on code task as well as iterative settings. In the table, the models for code and math tasks are trained separately. For math tasks, we still randomly sample from the Polaris dataset. For code tasks, our dataset is randomly sampled from AReaL-boba-2, and evaluation is conducted on LiveCodebench using the evalscope~\cite{evalscope_2024} framework.
For the code experiments, we observe that CE achieves relatively poor performance, even falling below the starting model. When using Strategy 2, we find that employing a higher learning rate for unlearned policy—specifically 5e-6 or 1e-5 instead of 1e-6—yields substantially better results. This aligns with our previous ablation study on signal construction, indicating that for unlearning strategies, the unlearning magnitude needs to be adjusted to provide an effective distillation signal.
For math tasks, we conduct three iterations on both Qwen3.5-9B and Qwen3.5-27B. We find that in the first iteration on Qwen3.5-9B, CE does not produce a sufficiently strong effect and fails to yield a very strong teacher. Instead, using MSE+Distillation leads to a stronger model. Across the three iterative rounds, both the 9B and 27B models show steady improvement. Notably, Qwen3.5-27B improves to a level comparable to some models with around 1 trillion total parameters, demonstrating the effectiveness of our method.
\textbf{A hyperparameter tuning suggestion}: when tuning the learning rate for the unlearned policy, we don’t necessarily need to train until recovery in every experiment, as that would take too long. Instead, we can pre-train a recovered policy $\pi_{\theta_e}$ using a fixed learning rate of 1e-6. Then, by adjusting the learning rate, we can additionally train a few other unlearned policies $\pi_{\theta_{\text{un}}}$ — for about ten steps — which will speed up the iteration process.

\begin{figure}[h]
    \centering
    \includegraphics[width=0.8\linewidth]{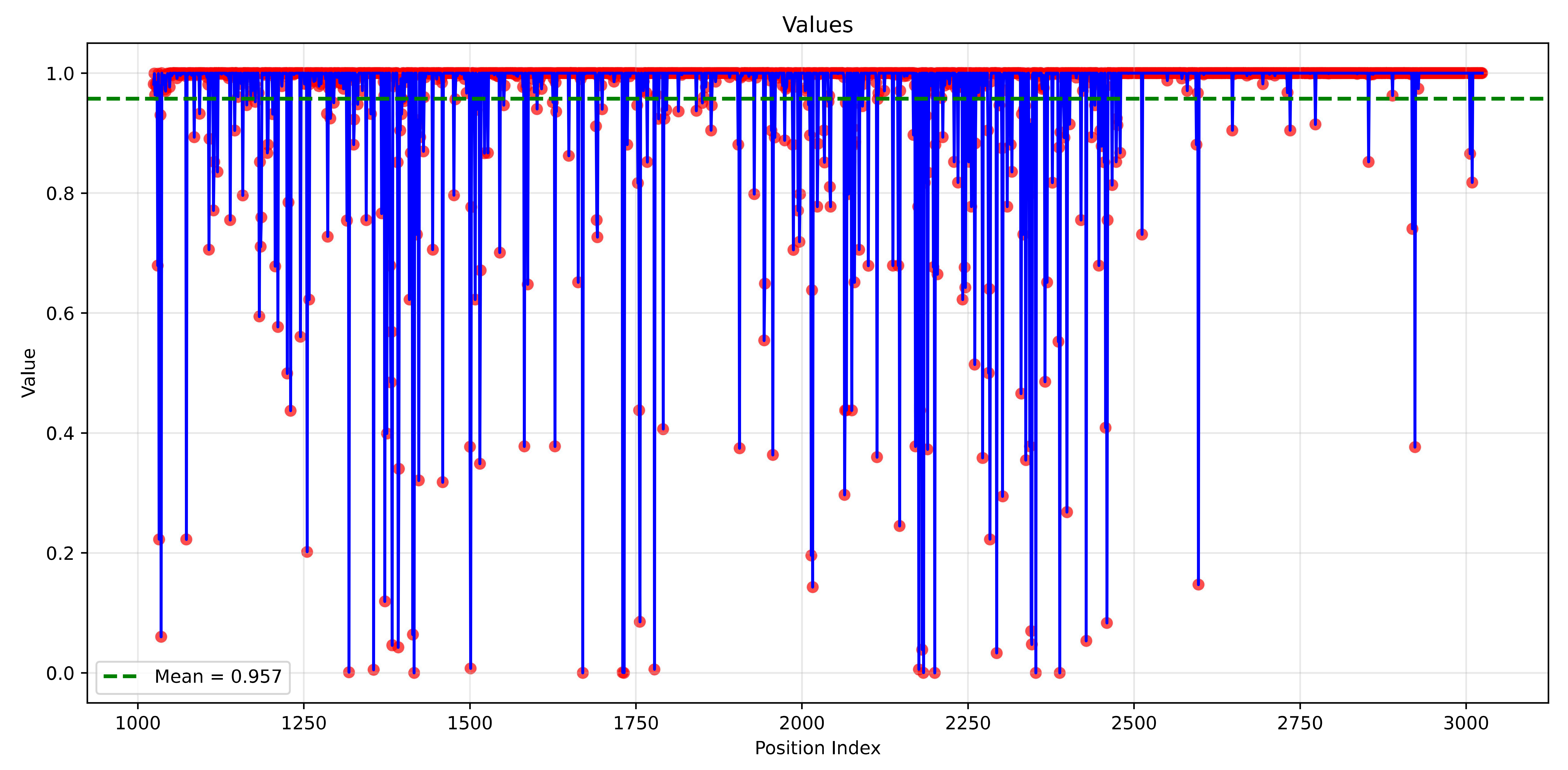}
    \caption{Although the probability of positive examples is close to one, there are quite a few outliers.}
    \label{fig:value}
\end{figure}

\textbf{Ablation on Outliers.} We observe that, for positive examples, the language-model policy head produces a number of probability outliers (Fig.~\ref{fig:value}), whereas negative examples remain smoothly distributed around zero.
In contrast, a conventional linear critic exhibits smoother behavior across positions.
We attribute this phenomenon to the softmax normalization in the policy head: since the total probability mass is constrained to sum to one, different actions compete when being pushed toward probability one.
Actions that lose this competition manifest as low-probability outliers.
We hypothesize that these outliers encode informative relative preferences, which we further examine through ablation studies.

\begin{table}[h]
\centering
\small
\caption{Ablation study on outliers in the MSE training strategy}
\setlength{\tabcolsep}{2pt}
\resizebox{0.6\linewidth}{!}{%
\begin{tabular}{llccc}
\toprule
\multirow{2}{*}{\textbf{Model}}
& \multirow{2}{*}{\textbf{Method}}
& \textbf{AIME} & \textbf{AIME} & \textbf{HMMT}  \\
&& \textbf{2024} & \textbf{2025} & \textbf{Feb 25}  \\
\midrule
\multirow{4}{*}{Qwen3-4B}
& Base & 84.9 & 81.1 & 56.0   \\
& \quad +Distillation (Original) & \underline{86.7} & \textbf{82.8} & \underline{60.1}    \\  
& \quad +Distillation ($\max(0, \cdot)$) & 86.6 & 81.4 & 59.5 \\  
& \quad +Distillation ($\min(0, \cdot)$) & \textbf{87.5} & \underline{81.9} & \textbf{62.7}    \\  
\midrule
\multirow{4}{*}{NBG4-3B}
& Base & 90.9 & 84.8 & 63.9   \\
& \quad +Distillation (Original) & \underline{91.2} & \underline{86.6} & \underline{67.6}   \\
& \quad +Distillation ($\max(0, \cdot)$) & 90.5 & 85.5 & 65.7\\
& \quad +Distillation ($\min(0, \cdot)$) & \textbf{91.3} & \textbf{87.4} & \textbf{68.6}  \\
\bottomrule     
\end{tabular}
}
\label{tab:out}
\end{table}

To examine whether these signals are meaningful, we conduct an ablation study
that selectively masks different regions of the log-ratio.
Specifically, for positive examples, we apply element-wise clipping to retain only
$\max(0, \cdot)$ or $\min(\cdot, 0)$ components of the log-ratio, thereby isolating
positive or negative signals, respectively.
In Tab.~\ref{tab:out}, we find that although negative log-ratio values (outliers) constitute only a small
fraction of all tokens (approximately 10\%), masking them leads to a noticeable
performance drop.
Moreover, retaining only these negative signals can outperform using both positive
and negative signals together. 
These results suggest that such outliers may not be purely noise; rather, they appear to encode informative corrective signals, potentially indicating actions that are disfavored by $\pi_{\theta_e}$ and warrant suppression during optimization.
However, the experiment in the table performed whitening after masking the signal, which may alter the overall distribution of the signal. In some experiments on Qwen3.5, removing these signals also does not lead to a significant drop in performance. As for what role outliers actually play, further experiments may be needed. Here, we only conduct preliminary experiments on outliers for reference, and we leave a thorough investigation of their role for future work.

%% file: iclr2026/src/related.tex
\section{Related works}

\textbf{Token-Level Signals for RL.} 
In reinforcement learning for large language models, a key challenge in improving sample efficiency is the sparsity of outcome-level rewards, which induces high-variance token-level optimization signals. To mitigate this issue, a widely studied line of work derives token-level rewards from the token probability distribution of language models.
A common approach uses the log-ratio
$
\log \frac{\pi_{\theta_\star}}{\pi_{\theta_{\mathrm{ref}}}}$
as a token-level reward, where $\pi^\star$ is typically a policy optimized by Direct Preference Optimization (DPO). \citet{rafailov2024r} show that, under a maximum-entropy RL framework, a DPO-optimized policy implicitly encodes $Q$-value information. Building on this insight, subsequent work leverages such token-level rewards to strengthen training signals across multiple algorithms, including additional rounds of DPO~\citep{zhutgdpo} and enhancements to PPO or RLOO~\citep{ce,cui2025process,RTO}. Among these methods, \citet{ce} directly parameterize a reward model using the log-ratio $\log \frac{\pi_\theta}{\pi_{\theta_{\mathrm{ref}}}}$ and fit it via a cross-entropy objective, which corresponds to the CE baseline we compare against.
In contrast, our work adopts a broader perspective on log-ratio-based token rewards. We study the interaction between token-level rewards and policy learning when trained on the same batch of data, including whether the student policy can surpass its teacher. We further examine the effect of more extreme MSE losses, providing empirical evidence linking critic-based methods~\citep{yue2025vapo,vcppo} with log-ratio-derived token rewards.

\textbf{Trade-off in RL Loss Design.} Prior work has designed a spectrum of variants ranging from conservative to aggressive updates for determining the appropriate loss function for a batch of data. The more conservative approaches include PPO and GRPO~\citep{ppo,grpo}, which employ a clipping function that zeroes out the gradient once the discrepancy between the old and new policies exceeds a certain threshold. Subsequent methods have relaxed the gradient update magnitude to varying degrees. For instance, SAPO~\citep{sapo} replaces the hard truncation of the clipping function with a soft truncation via the sigmoid function. VESPO~\citep{vespo} also adopts soft truncation; however, empirically it proves more conservative than SAPO, assigning extremely small gradients in regions of large discrepancy, thereby enabling safer execution of multiple off-policy steps. The most aggressive methods, such as CISPO~\citep{cispo} and DISPO~\citep{dispo}, follow the REINFORCE algorithm without any gradient truncation. Such unclipped algorithms are susceptible to the influence of negative gradients and are more prone to excessive policy deviation. As demonstrated by our experiments, soft-truncation methods like SAPO are more sample-efficient than hard-clipping methods like GRPO. Fundamentally, this family of algorithms addresses the trade-off between update magnitude and sample efficiency. Our method decouples these two concerns, optimizing them in separate stages. In this paper, we leverage the proximity to the initial model to filter out components irrelevant to model improvement; future work may explore more refined techniques for impurity removal during the distillation process.

\section{Conclusion}
We have shown that the trade-off between sample efficiency and optimization stability can be addressed by decoupling. Instead of forcing a single policy to balance aggressive updates and stable training, our two-stage framework first extracts rich training signals through weakly constrained off-policy optimization, then distills these signals into a KL-constrained policy. This decoupling allows each stage to focus on its own objective without compromising the other.
Several directions remain for future work. First, our weak-to-strong results show that even suboptimal teachers can provide useful signals, but we currently lack a clear theoretical explanation for when and why this happens. A deeper understanding of this phenomenon could help identify the best conditions for weak teacher distillation. Second, while the second stage uses KL constraints to filter out spurious policy drift, our experiments also suggest that entropy reduction may be another source of noise unrelated to actual task improvement. A more systematic analysis of such noise components and the design of corresponding denoising methods could further improve distillation quality.